\documentclass[10pt,twocolumn,letterpaper]{article}

\usepackage{iccv}
\usepackage{times}
\usepackage{epsfig}
\usepackage{graphicx}
\usepackage{amsmath}
\usepackage{amssymb}

\usepackage{times}
\usepackage{epsfig}
\usepackage{graphicx}
\usepackage{amsmath}
\usepackage{amssymb}

\usepackage{enumerate}
\usepackage{graphicx}
\usepackage{subcaption}
\usepackage[shortlabels]{enumitem}
\usepackage{textcomp}
\usepackage{authblk}
\usepackage{makecell}
\usepackage[pagebackref=true,breaklinks=true,letterpaper=true,colorlinks,bookmarks=false]{hyperref}

\iccvfinalcopy % *** Uncomment this line for the final submission

\newcommand{\eqnref}[1]{Eq.~\ref{#1}}
\newcommand{\figref}[1]{Fig.~\ref{#1}}

\makeatletter
\renewcommand\AB@affilsepx{, \protect\Affilfont}
\makeatother

\def\sec#1{Sec.~\ref{#1}}

\ificcvfinal\fi

\begin{document}

%%%%%%%%% TITLE
\title{Learning Efficient Photometric Feature Transform for Multi-view Stereo}
\author[1]{Kaizhang Kang}
\author[1]{Cihui Xie}
\author[1]{Ruisheng Zhu}
\author[1]{Xiaohe Ma}
\author[2]{Ping Tan}
\author[1]{Hongzhi Wu\thanks{: corresponding author (hwu@acm.org).}}
\author[1]{Kun Zhou}
\affil[1]{Zhejiang University}
\affil[2]{Simon Fraser University}

\maketitle
\ificcvfinal\thispagestyle{empty}\fi

%%%%%%%%% ABSTRACT
\begin{abstract}

We present a novel framework to learn to convert the per-pixel photometric information at each view into spatially distinctive and view-invariant low-level features, which can be plugged into existing multi-view stereo pipeline for enhanced 3D reconstruction. Both the illumination conditions during acquisition and the subsequent per-pixel feature transform can be jointly optimized in a differentiable fashion. Our framework automatically adapts to and makes efficient use of the geometric information available in different forms of input data. High-quality 3D reconstructions of a variety of challenging objects are demonstrated on the data captured with an illumination multiplexing device, as well as a point light. Our results compare favorably with state-of-the-art techniques.

\end{abstract}

%%%%%%%%% BODY TEXT
\section{Introduction}

As one central problem in computer vision and graphics, shape reconstruction in the presence of complex appearance is challenging. At one hand, multi-view stereo methods~\cite{furukawa2015multi} usually require a Lambertian-dominant reflectance for computing reliable view-invariant features. The appearance variation with view or lighting is undesirable, as it may change the native spatial features on the object, or specularly reflect the projected pattern from active illumination~\cite{levoy2000digital,salvi2004pattern}, leading to difficulties in correspondence computation for shape reconstruction.

On the other hand, single-view photometric stereo~\cite{woodham1980photometric,chen2018ps} exploits the lighting variations on each pixel and transforms the image measurements into a normal map. While high-quality details can be recovered, it suffers from low-frequency shape distortions~\cite{nehab2005efficiently}. Recently, multi-view photometric stereo~\cite{hernandez2008multiview,vlasic2009dynamic,li2020multi} accurately integrates the photometric cues from different viewpoints, subject to the geometric constraints across multiple views. 

However, photometric stereo techniques are not scalable to the amount of geometric information in the input data for 3D reconstruction, leading to suboptimal results. When additional physical cues such as rapid albedo variations are present, they cannot be exploited to improve the reconstruction quality. On the other hand, when the measured photometric information is insufficient to determine a normal field, the quality of results from existing approaches will significantly degrade. Furthermore, related techniques heavily exploit reflectance properties~\cite{li2020multi,logothetis2019differential}, which hinders the extension to handle more general appearance such as anisotropic materials.

To tackle the above challenges, we make the key observation that it is not necessary to use \textbf{normal} as the intermediate representation for 3D reconstruction from photometric measurements. Instead, we propose a novel differentiable framework, to efficiently transform the per-pixel photometric information measured at each view, into \textbf{automatically learned low-level features}, in an end-to-end fashion. The learned per-pixel features essentially exploit the available geometric information in photometric measurements, and can be plugged in existing multi-view stereo pipelines for further processing tasks like spatial aggregation, resulting in enhanced geometric reconstruction. Furthermore, our data-driven framework is highly flexible and can adapt to various factors, including the physical acquisition capabilities / characteristics of different setups, and different types of appearance such as anisotropic reflectance, by feeding corresponding training data.

The effectiveness of our framework is demonstrated with a high-performance illumination multiplexing device, on  geometric reconstruction results of a variety of 3D objects using as few as 16 input photographs per view. Moreover, the framework is generalized to handle the input data of conventional photometric stereo with one point light on at a time (DiLiGenT-MV\cite{li2020multi}). Our results compare favorably with state-of-the-art techniques. We make public the code and data of this project at \emph{\$URL to be revealed upon acceptance\$}.

\section{Related work}

\subsection{Multi-view Stereo} 
These methods~\cite{furukawa2015multi} first compute low-level features in multi-view input images, which ideally should be discriminative in the spatial domain, and invariant with respect to various factors. Next, the correspondences between features at different views are established, and used to determine 3D points via triangulation. Since the raw measurements at a single pixel are usually not sufficient to accurately establish multi-view correspondences, spatial aggregation is typically performed to incorporate more information from neighboring pixels.

Excellent results with Lambertian-dominant materials are demonstrated from photographs taken with even uncontrolled conditions~\cite{galliani2015massively,schoenberger2016sfm,schoenberger2016mvs}. But these methods fail on textureless objects, as the features across different locations may be almost identical. Active lighting methods, such as laser-stripe triangulation~\cite{levoy2000digital} or structured lighting~\cite{salvi2004pattern}, handle such cases by physically projecting spatially distinctive patterns onto the object. Additional physical dimensions like polarization can also be exploited~\cite{cui2017polarimetric}. Recently, machine learning further pushes the reconstruction quality, by replacing hand-crafted features with automatically learned ones~\cite{simonyan2014learning,zbontar2015computing,zagoruyko2015learning,tian2017l2}. However, for complex appearance that changes with view / lighting conditions, it is still challenging to compute distinctive and invariant features for 3D reconstruction.

Our work is orthogonal to the majority of work here, which focuses on processing information in the spatial domain. Instead, we focus on the angular domain, by learning to transform the per-pixel photometric information into useful features. We employ one existing pipeline for subsequent spatial-domain processing, and leave the unified treatment of both domains for future work. It is worth mentioning that the work of~\cite{wu2018specular} transforms multi-view images of a homogeneous isotropic reflectance into a diffuse one, to reduce the view variance of specular reflections.

\subsection{Photometric Stereo}
This class of techniques compute an accurate normal field that can be subsequently integrated into a depth map, from appearance variations under typically a large number of different illumination conditions~\cite{shi2016benchmark}. Starting from the seminal work of~\cite{woodham1980photometric} with a Lambertian reflectance and a calibrated directional light, substantial efforts been made to extend to handle more general materials~\cite{alldrin2008photometric,goldman2009shape,shi2012elevation} and/or unknown lighting conditions~\cite{alldrin2007resolving,basri2007photometric,lu2013uncalibrated}.

The closest work to ours is multi-view photometric stereo. While traditional approaches consider a single view only, it combines photometric cues at multiple different views to produce a complete 3D shape. An initial coarse geometry is refined with the normal information, as in~\cite{hernandez2008multiview,zhou2013multi,li2020multi}. The depth maps integrated from normal fields are directly fused to obtain the final result, according to~\cite{vlasic2009dynamic}. And the relationship between a signed distance field and normals is exploited in~\cite{logothetis2019differential}.

Our work also takes as input photometric information for 3D reconstruction. The main difference is that we do not use a fixed intermediate representation (e.g., normal), and instead compute learned low-level features that efficiently scale to the amount of geometric information available in the input data. Moreover, inspired by recent work on differentiable reflectance capture~\cite{Kang:2019:JOINT}, our physical acquisition process can be jointly optimized with the computational transformation of features.

\section{Acquisition Setup} 
\begin{figure}
   \centering
      \includegraphics[height = 1.6in]{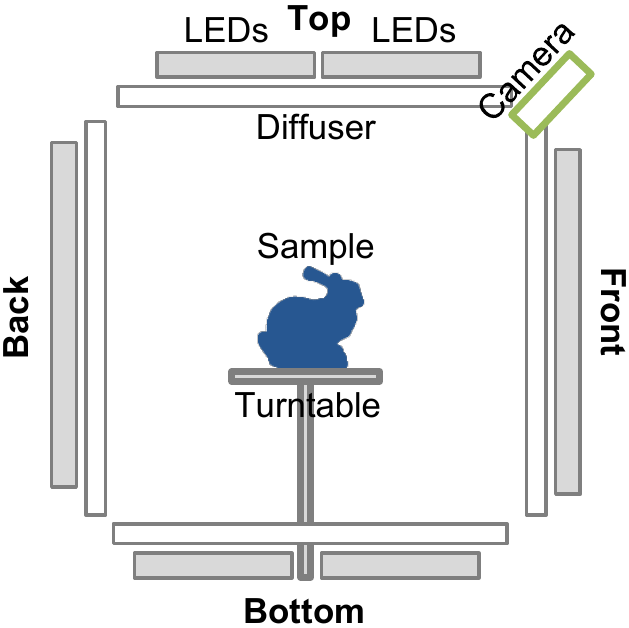}
      \hspace{0.1in}
      \includegraphics[height = 1.6in]{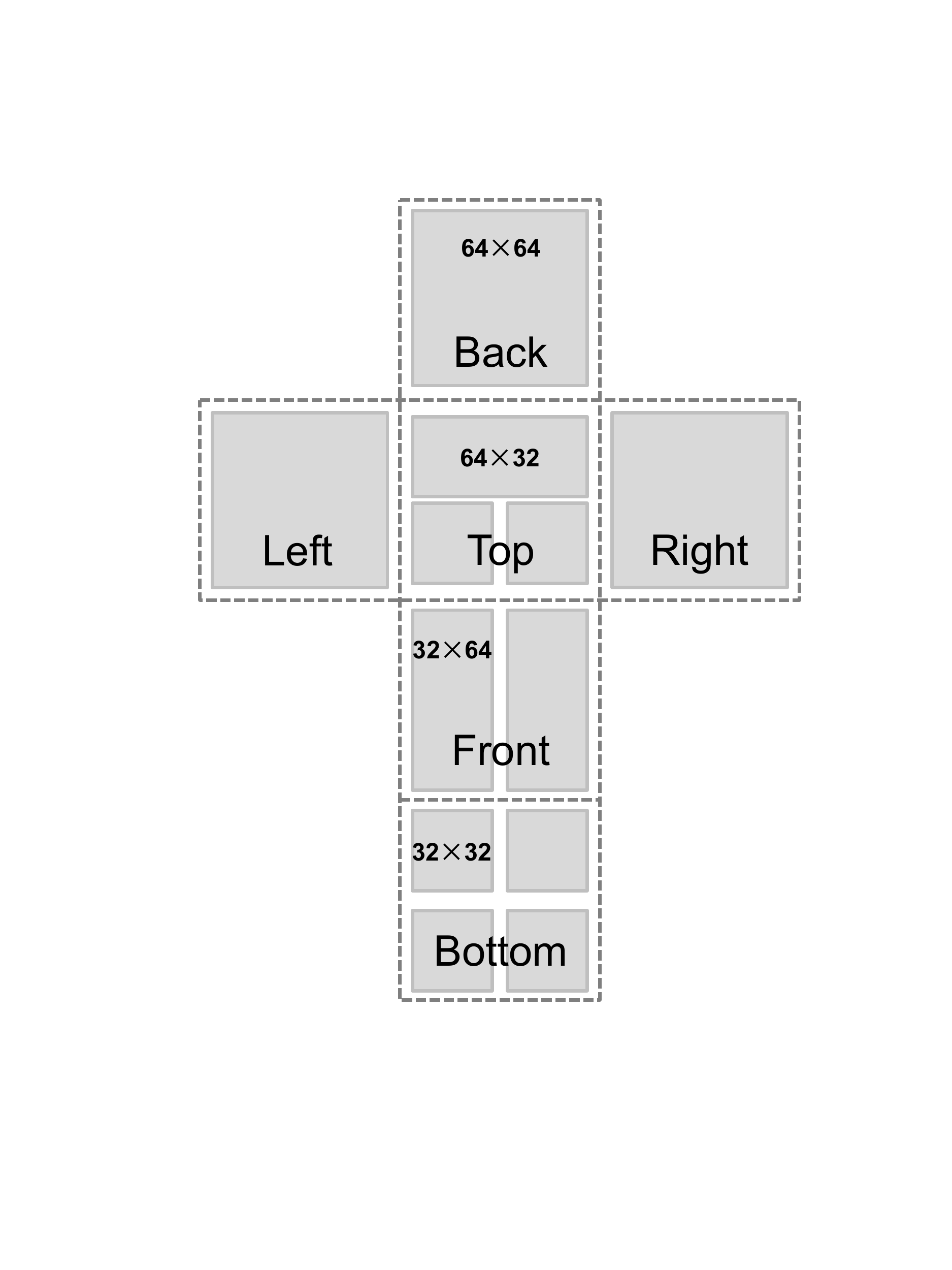}
  	\caption{The lighting layout of our setup. A side view of the setup (left) and the vertical-cross parameterization of all lights with 4,096 LEDs on each face (right).}
   \label{fig:device}
\end{figure}

Our main experiments are conducted with a high-performance, box-shaped lightstage, similar to~\cite{Kang:2019:JOINT}. Its size is 80cm $\times$ 80cm $\times$ 77cm. The sample object is placed on a digital turntable near the center of the device, and rotated to different angles for multi-view imaging. A single FLIR BFS-U3-123S6C-C vision camera captures photographs at a resolution of 4,096$\times$3,000. We illuminate the sample with 24,576 LEDs on the six faces of the device with polycarbonate diffusers attached. The total LED power is about 2,000W, and the pitch of adjacent LEDs is 1cm. We calibrate the intrinsic / extrinsic parameters of the camera, as well as the positions, orientations, angular intensity of LEDs. The rotation angle of the turntable is computed from printed markers on its surface~\cite{1467495}. Please refer to~\figref{fig:device} for an illustration.
\begin{figure*}[htb]
\centering
	\includegraphics[width = \linewidth]{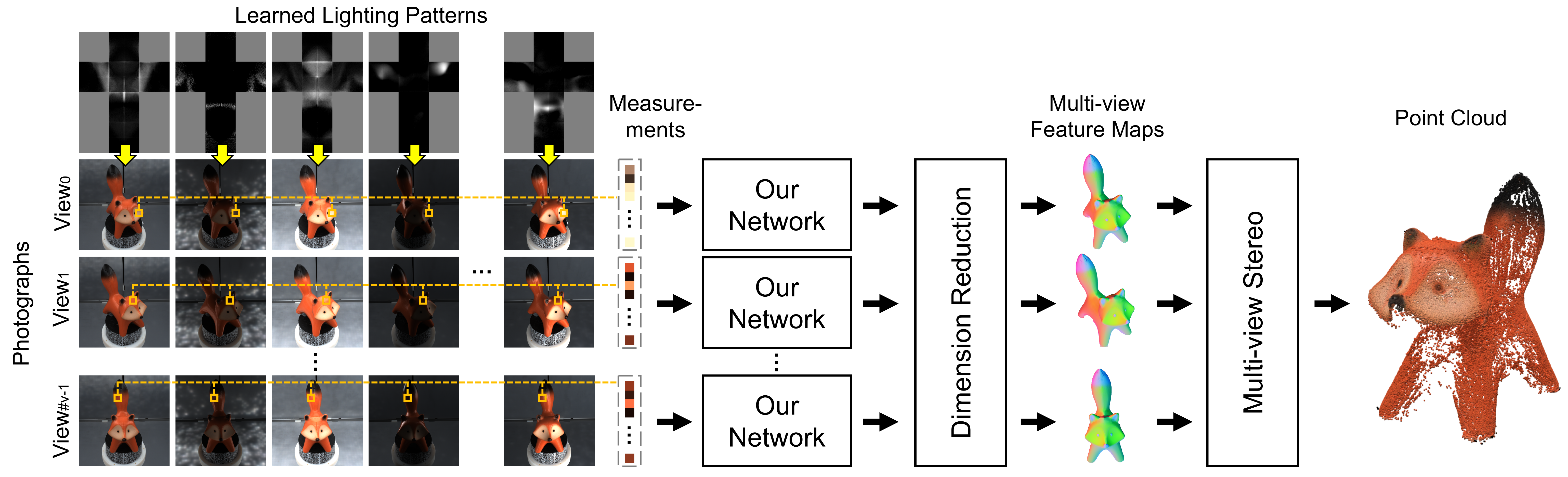}
    	\caption{Our pipeline. First, we take photographs of a 3D object under learned lighting patterns at multiple views. For \textbf{each pixel} at each view, our network transforms its photometric measurements to a high-dimensional feature vector. Next, dimension reduction is performed on all transformed features across different locations and views, resulting in multi-view feature maps. Finally, we send these maps as input to a multi-view stereo technique for geometric reconstruction.}
  \label{fig:pipeline}
\end{figure*}
\section{Preliminaries} 
The following derivations are based on a  gray-scale channel. First, the outgoing radiance $B$ from a surface point $\mathbf{p}$ towards a camera can be modeled as~\cite{pharr2016physically}:
\begin{align}
B(I; \mathbf{p}) =  &\sum_{l}  I(l)  \int  \frac{1}{|| \mathbf{x_{l}} - \mathbf{x_{p}} ||^2} \Psi(\mathbf{x_{l}}, -\mathbf{\omega_{i}}) V(\mathbf{x_{l}}, \mathbf{x_{p}})  \nonumber \\
& f(\mathbf{\omega_{i}}'; \mathbf{\omega_{o}}', \mathbf{p}) (\mathbf{\omega_{i}} \cdot \mathbf{n_{p}})^{+} (-\mathbf{\omega_{i}} \cdot \mathbf{n_{l}})^{+} d\mathbf{x_{l}}.
\label{eq:render}
\end{align}
Here each light $l$ is a locally planar source. $\mathbf{x_{p}}$~/~$\mathbf{n_{p}}$ is the position~/~normal of $\mathbf{p}$, while $\mathbf{x_{l}}$~/~$\mathbf{n_{l}}$ is the position~/~normal of a point on the light source $l$. $\mathbf{\omega_{i}}$~/~$\mathbf{\omega_{o}}$ denotes the lighting~/~view direction in the world space, and $\mathbf{\omega_{i}}'$~/~$\mathbf{\omega_{o}}'$ is expressed in the local frame of $\mathbf{p}$. Note that $\mathbf{\omega_{i}} = \frac{\mathbf{x_{l}} - \mathbf{x_{p}}}{|| \mathbf{x_{l}} - \mathbf{x_{p}} ||}$. $I(l)$ is the intensity for the light $l$, in the range of [0, 1]. The array $\{ I(l) \}$ corresponds to a \textbf{lighting pattern}. $\Psi(\mathbf{x_{l}}, \cdot)$ represents the angular distribution of the light intensity. $V$ is a binary visibility function between $\mathbf{x_{l}}$ and $\mathbf{x_{p}}$. The operator $( \cdot )^{+}$ computes the dot product between two vectors, and clamps any negative result to zero. $f(\cdot;  \mathbf{\omega_{o}}', \mathbf{p})$ is a 2D BRDF slice, which is a function of the lighting direction. In this paper, we employ a standard anisotropic GGX model~\cite{10.5555/2383847.2383874} to represent $f$. Note that our framework is not tied to the choice of the BRDF model.

As $B$ is linear with respect to $I$ (\eqnref{eq:render}), it can be expressed as the dot product between $I$ and a \textbf{lumitexel} $c$:
\begin{equation}
  B(I; \mathbf{p}) = \sum_{l} I(l) c(l; \mathbf{p}),
\label{eq:linear}
\end{equation}
where $c$ is a function of the light source $l$, defined on the surface point $\textbf{p}$ of the sample object:
\begin{equation}
c(l; \mathbf{p}) = B(\{ I(l) = 1, \forall_{j \neq l } I(j) = 0 \}; \mathbf{p}).
\label{eq:lumitexel}
\end{equation}
Each element of $c$ records a measurement $B$ with one light on at a time. 

\section{Overview}
We propose a mixed-domain neural network, to physically encode the photometric information into a small number of measurements by projecting learned lighting patterns, and then computationally transform spatially discriminative and view/lighting-invariant low-level features, on a per-pixel basis. The procedure is repeated for each pixel in each image taken at a particular view, resulting in multi-view, high-dimensional transformed feature maps. These maps are then post-processed and fed into an existing multi-view stereo technique to produce the final 3D geometry. \figref{fig:pipeline} illustrates the process.

While the network is originally designed to work in conjunction with our lightstage, we will describe how to generalize to handle the input of conventional photometric stereo in~ \sec{sec:details}. Note that unlike related work on learning-based features~\cite{simonyan2014learning,zagoruyko2015learning,tian2017l2}, we choose not to perform spatial aggregation in the network, due to the lack of high-quality, large-scale databases of 3D objects with complex spatially-varying appearance. 

\section{Our Network}

\subsection{Input/Output}
The main input to the network is a physical grayscale lumitexel at a 3D surface point corresponding to a particular pixel, representing the maximum amount of information one can acquire with an illumination multiplexing setup. The parameterization of a lumitexel is the same as the lighting layout (\figref{fig:device}). 
The other input is the view specification of the current image, represented as $[\cos(\theta),\sin(\theta)]$ for continuity, in which $\theta$ is the rotation angle of the turntable along its axis. The purpose of adding this extra input is to guide the network to learn global features that are view invariant. The output is a high-dimensional feature vector. The extension to handle RGB input is described in~\sec{sec:details}.

\subsection{Architecture}
\label{subsec:net_arch}
\begin{figure*}[htb]
\centering
	\includegraphics[width = \linewidth]{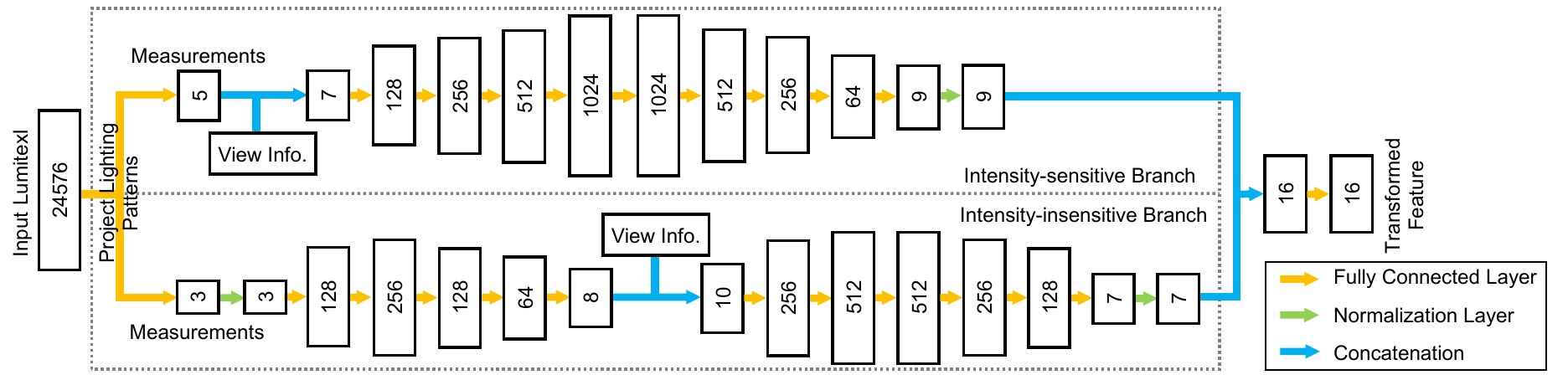}
	\vspace{-0.1in}
    	\caption{The architecture of our neural network, consisting of a measurement-intensity-sensitive branch and an insensitive one. For each branch, the input physical lumitexel is converted into a small number of measurements, with a linear fc layer that corresponds to the lighting patterns used in acquisition. The measurements, along with the view information of the current photograph, are transformed via a series of non-linear fc layers, before normalized to a unit feature vector. Finally, a linear fc layer combines the results from the two branches to output the final feature.}
  \label{fig:network}
\end{figure*}

Our network consists of two main branches that are designed to exploit different aspects of the input photometric information. Each branch produces a separate feature vector. The two vectors are then combined with a linear fully-connected (fc) layer to generate the final feature. A graphical illustration is shown in~\figref{fig:network}.

The first branch is an intensity-sensitive one. Its first layer is a linear fc one, whose weights correspond to the lighting patterns during acquisition. This layer encodes the physical lumitexel into a small number of measurements. These measurements, along with the view specification, then go through 9 fc layers. Finally, a normalization layer produces a unit feature vector as output for training stability, as common in feature learning literature~\cite{schroff2015facenet,wu2017sampling}.

The other branch is designed to be insensitive to the intensity of measurements. Overall its structure is similar to the first branch, with two major exceptions. First, there is an additional normalization layer right succeeding the measurements, to encourage the network to focus on photometric cues that are not related to input intensities. Second, the input view specification is added 5 layers after the first normalization one, to allow sufficient processing of the measurements.

\begin{figure}
   \begin{minipage}{\linewidth}
      \centering
      \includegraphics[width=0.3\linewidth]{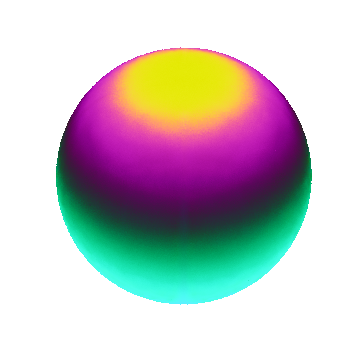}
      \includegraphics[width=0.3\linewidth]{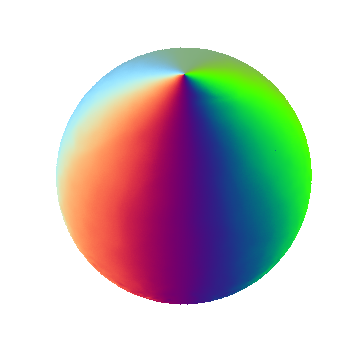}
      \includegraphics[width=0.3\linewidth]{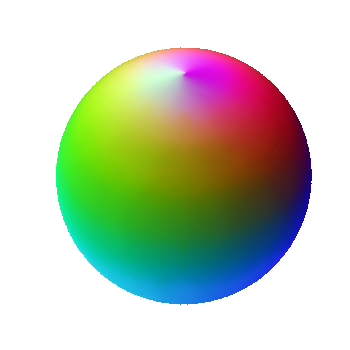}
   \end{minipage}
   \caption{Visualization of learned features, computed on a synthetic ball with a homogeneous BRDF. For visualization, the original high-dimensional features are first projected to 3D via PCA and then adjusted to the range of $[0, 1]$ on a per-channel basis. From the left to right, the features from the intensity-sensitive / insensitive branch, and our network that combines the two.}
   \label{fig:feature_space}
\end{figure}
Our initial attempt is to use the intensity-sensitive branch only, as it is a straight-forward, end-to-end architecture. However, the resultant features hardly exhibit variations along the azimuth angle, as visualized in~\figref{fig:feature_space}. Adding the intensity-insensitive branch helps increase the discrimination power of features (\figref{fig:feature_space}) and reduce the network loss at the same time (\sec{sec:eval}).

Note that to physical realize the lighting patterns and to prevent degradation in the presence of training noise (\sec{sec:training}), we normalize the weights in the first fc layer of each branch that correspond to a pattern, similar to~\cite{Kang:2019:JOINT}. During acquisition, each pattern is split into a positive and a negative one, and scaled by the inverse of the weight of the maximum absolute value, to fit in the range of $[0,1]$.

\subsection{Loss Function}
\label{subsec:loss_training}
Ideally, features for multi-view stereo should have the following properties:
(1) the features of the same 3D point at different views are invariant;
(2) the features of two different points are sufficiently discriminative;
(3) and it is efficient to compare two features for matching.

Towards these goals, we define the following loss function to automatically learn to generate features embedded in a Euclidean latent space:
\begin{equation}
L = L_{\operatorname{main}} +\lambda  L_{\operatorname{reg}}.
\end{equation}
Here $L_{\operatorname{main}}$ is the term to enforce the distinctiveness and invariance of features. It is modified from the $E_1$ term proposed in L2-net~\cite{tian2017l2} as:
\begin{equation}
L_{\operatorname{main}} = -\frac{1}{2}\left(\sum_a\log{s^{\operatorname{col}}_{aa}}+\sum_a\log{s^{\operatorname{row}}_{aa}}\right),
\end{equation} 
where the related terms are defined as:
\begin{equation}
s^{\operatorname{row}}_{ij} = \frac{\exp\left(d_{ij}\right)}{\sum_{a=1}^k \exp\left(d_{ia}\right)}, \;\;
s^{\operatorname{col}}_{ij} = \frac{\exp\left(d_{ij}\right)}{\sum_{a=1}^k \exp\left(d_{aj}\right)}.
\label{eq:relative}
\end{equation}
Note that $d_{ij}$ is an entry in a feature distance matrix, constructed as follows. For each of $k$ training points, we randomly sample related generative parameters (\sec{sec:training}), and two visible view specifications. Then we compute the corresponding lumitexels at the two sampled views for each point according to~\eqnref{eq:lumitexel}, resulting in $\{ c_1^1, c_1^2, c_2^1, c_2^2, ..., c_k^1, c_k^2 \} $. Here the superscript denotes the view, while the subscript represents the point. Next, we transform these lumitexels using the current network into a set of feature vectors, as  $\{ h_1^1, h_1^2, h_2^1, h_2^2, ..., h_k^1, h_k^2 \} $. Following~\cite{tian2017l2},  we define $d_{ij} = || h_i^1 - h_j^2 ||_2$. This absolute Euclidean distance is converted to a relative one in~\eqnref{eq:relative}. Note that by minimizing $L_{\operatorname{main}}$, we are essentially maximizing $d_{ij}$ for $i \neq j$ and minimizing $d_{ii}$, which correspond to the distinctiveness and invariance properties respectively.

We add a regularization term $L_{\operatorname{reg}}=\sum_w \left\|w\right\|_2$ to the last layer to avoid overfitting, where $w$ is a weight in that layer. We use $\lambda = 3$ in all experiments.

\subsection{Training}
\label{sec:training}
Similar to previous work on differential reflectance acquisition~\cite{Kang:2018:ERC:3197517.3201279,Kang:2019:JOINT},
we synthesize the training lumitexels by randomly sampling parameters for evaluating~\eqnref{eq:lumitexel} based on the GGX BRDF model, including the geometric parameters (the 3D position and the local frame), the material parameters (diffuse/specular albedos and anisotropic roughnesses), and a view specification $\theta$. We use $\theta$ to transform the sampled position and local frame, which simulates the rotation of the turntable during acquisition. Also, we multiply each measurement with a Gaussian noise ($\mu = 1$, $\sigma = 1\%$)  during training, to increase the robustness of the network.

\section{Implementation Details}
\label{sec:details}

To handle RGB input photographs, for each pixel we apply the network to its measurements at each channel, and concatenate the per-channel results as a long vector. For efficient subsequent processing, dimensional reduction via principal component analysis (PCA) is performed on all transformed features across different views. The reduced dimensional feature maps are then fed to our modified version of COLMAP~\cite{schoenberger2016sfm,schoenberger2016mvs}, which operates on multi-channel floating-point input images.

To generalize to the input data in one state-of-the-art multi-view photometric stereo method~\cite{li2020multi}, we replace the input of the network with a vector that concatenating all 96 measurements under different lighting conditions. This vector will be sent as input to both the intensity-sensitive and insensitive branches. As a result, the original fc layer that represents the lighting patterns (\figref{fig:network}) is removed.

\section{Experiments}
Our network is implemented with PyTorch~\cite{NEURIPS2019_9015} on a workstation with an Intel Core i9-10940X CPU, 256GB memory and a GeForce GTX 2080 Ti graphics card. We first pre-train each branch for 100K iterations in an hour. After that, we train the complete network for 300K iterations in about 4 hours. The learning rate is set to $10^{-4}$. For each view, we use a total number of 8 photometric measurements, with 5 for the intensity-insensitive branch and 3 for the sensitive one. This corresponds to $8\times2=16$ non-negative lighting patterns / input photographs per view, which takes 15 seconds to capture with our setup. To scan a 3D object, we rotate the turntable to 24 equally spaced angles ($\theta_{i} = \frac{i}{24} 2\pi $, $i = 0, 1, ..., 23$). We use a feature length of 16 for each color channel. The dimension is reduced to 4 with PCA before sending to COLMAP. It takes 1.3 second to transform 1 million features using the network, and 1.5 hours for our modified COLMAP to compute a dense 3D point cloud out of transformed feature maps. 

\begin{figure}[htb]
      
      \begin{minipage}{0.05in}	
	    	   \centering
        		\rotatebox{90}{\small Photographs}
		      %\hspace{-0.1in}
      \end{minipage}
      \hspace{0.01in}	
      \begin{minipage}{3.1in}
         % \begin{subfigure}[b]{\linewidth}
            \centering
            \includegraphics[width=0.24\linewidth]{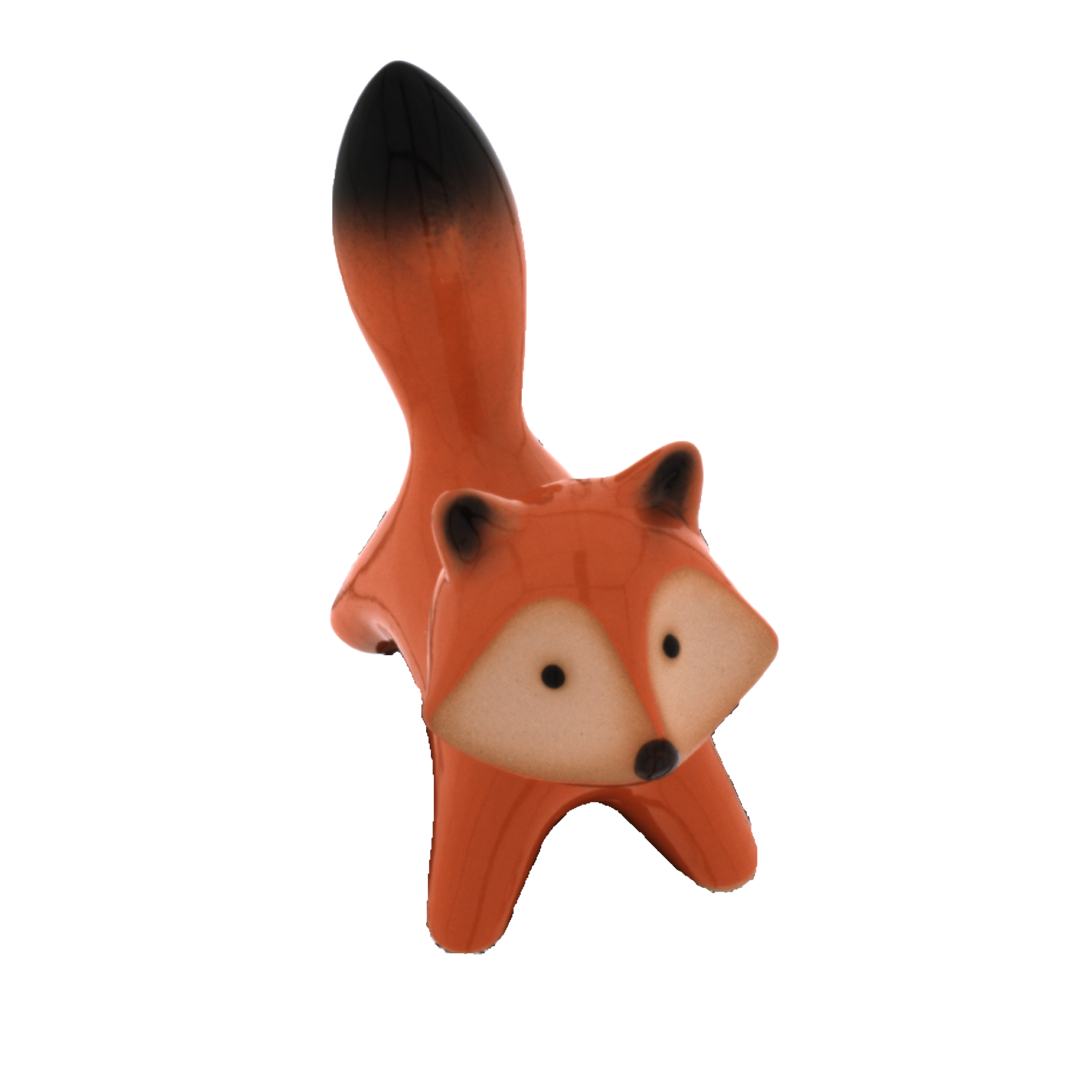}
            \includegraphics[width=0.24\linewidth]{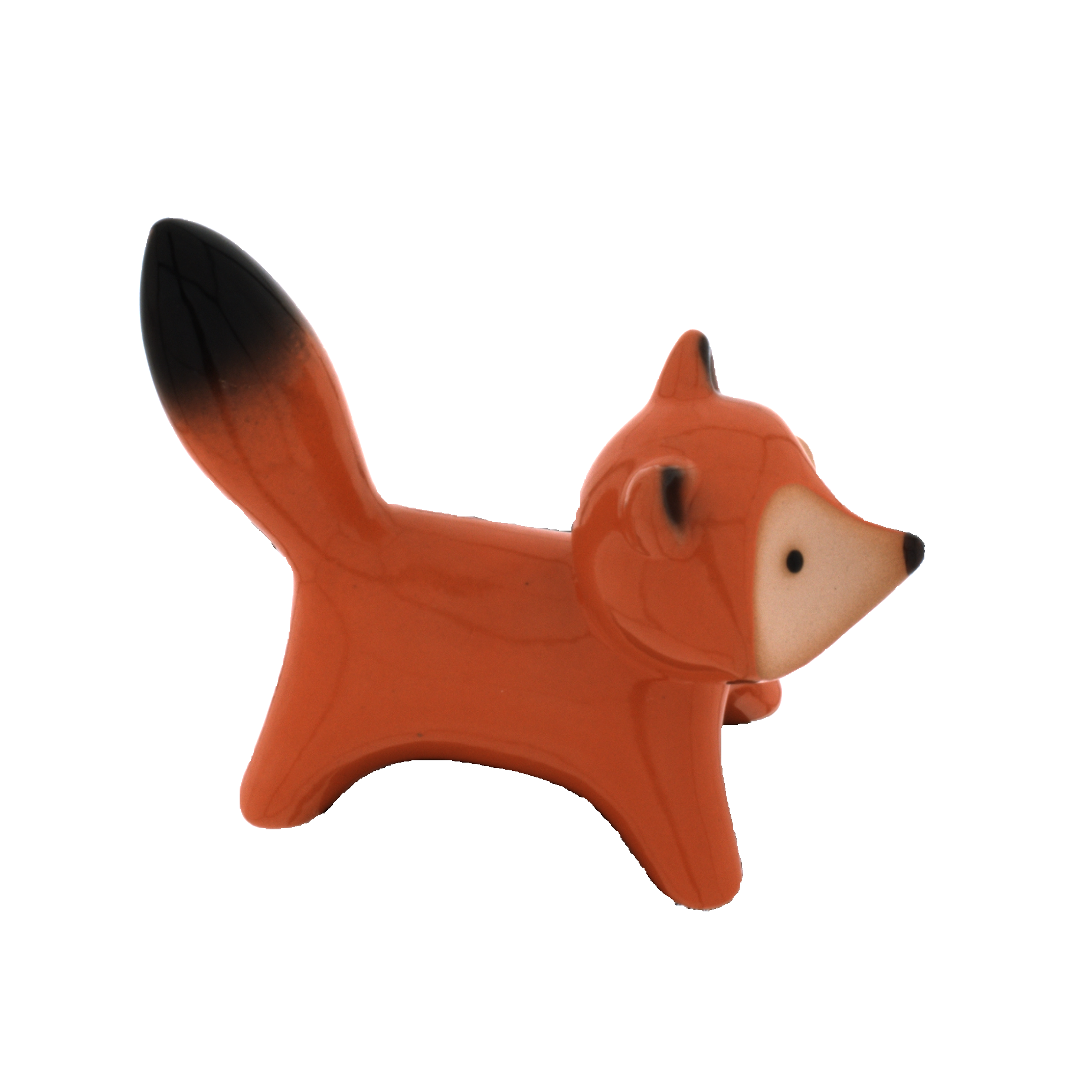}
            \includegraphics[width=0.24\linewidth]{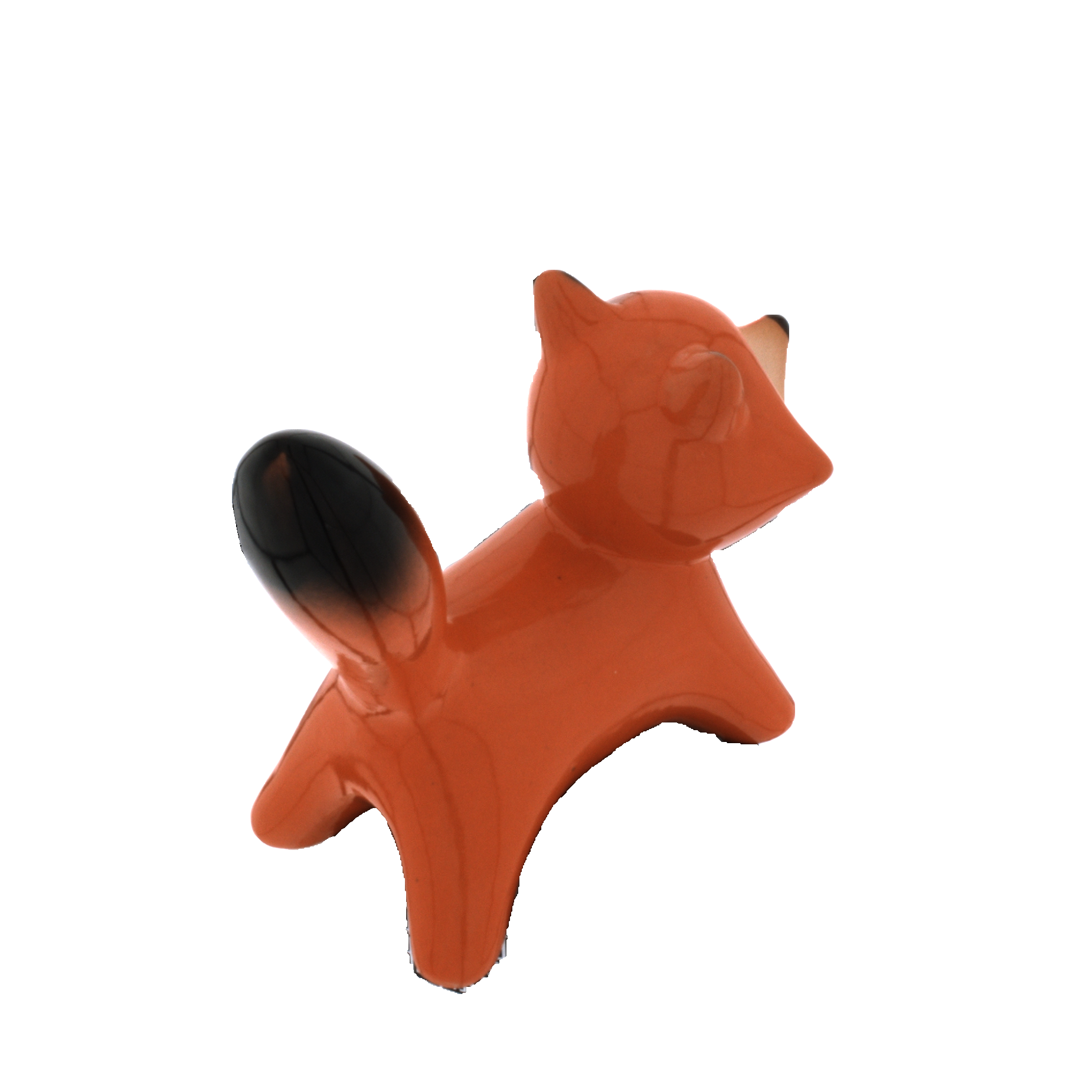}
            \includegraphics[width=0.24\linewidth]{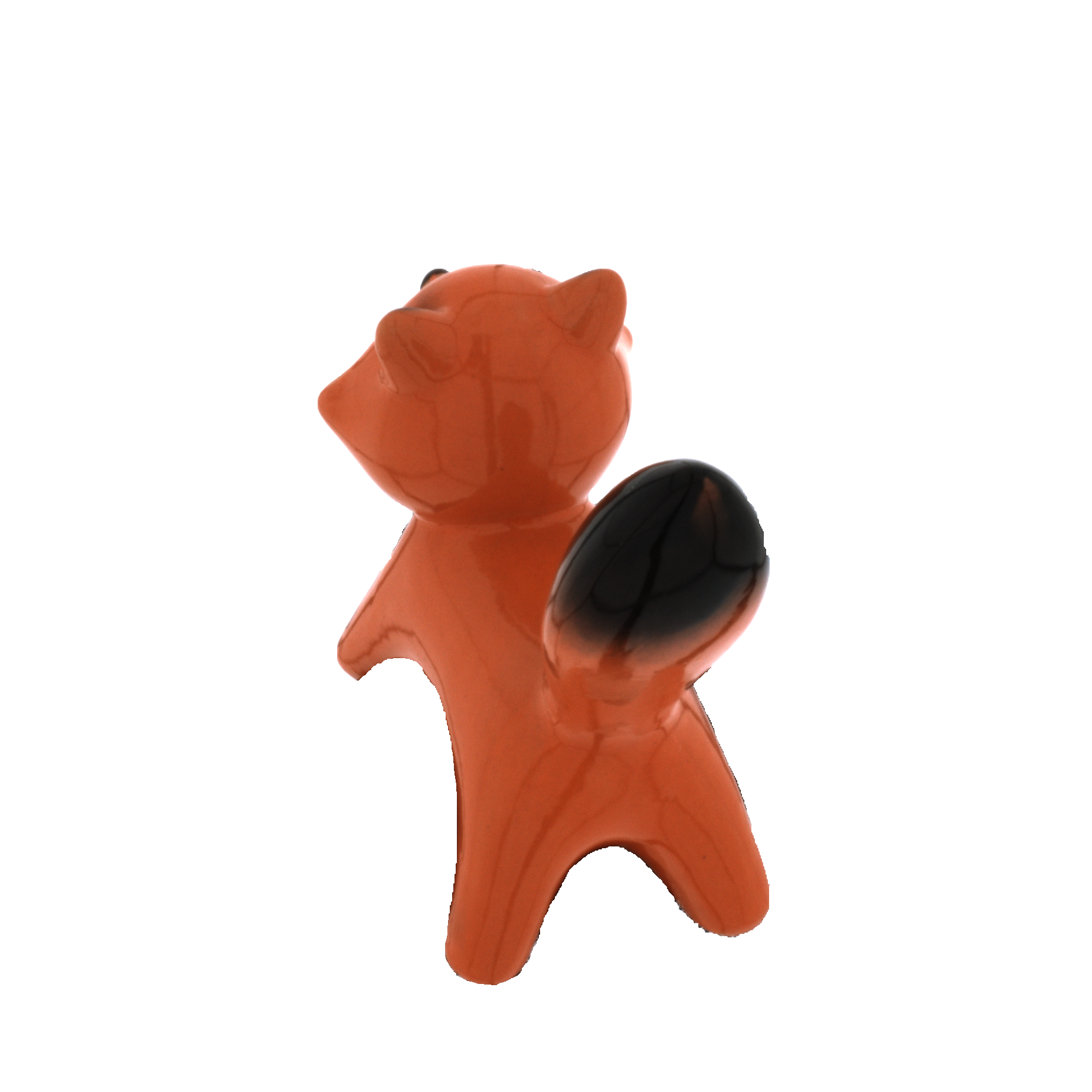}
       \end{minipage}
      
      \begin{minipage}{0.05in}	
	    	   \centering
        		\rotatebox{90}{\small L2-Net~\cite{tian2017l2}}
      \end{minipage}	
      \hspace{0.01in}	
      \begin{minipage}{3.1in}
         % \begin{subfigure}[b]{\linewidth}
            \centering
            \includegraphics[width=0.24\linewidth]{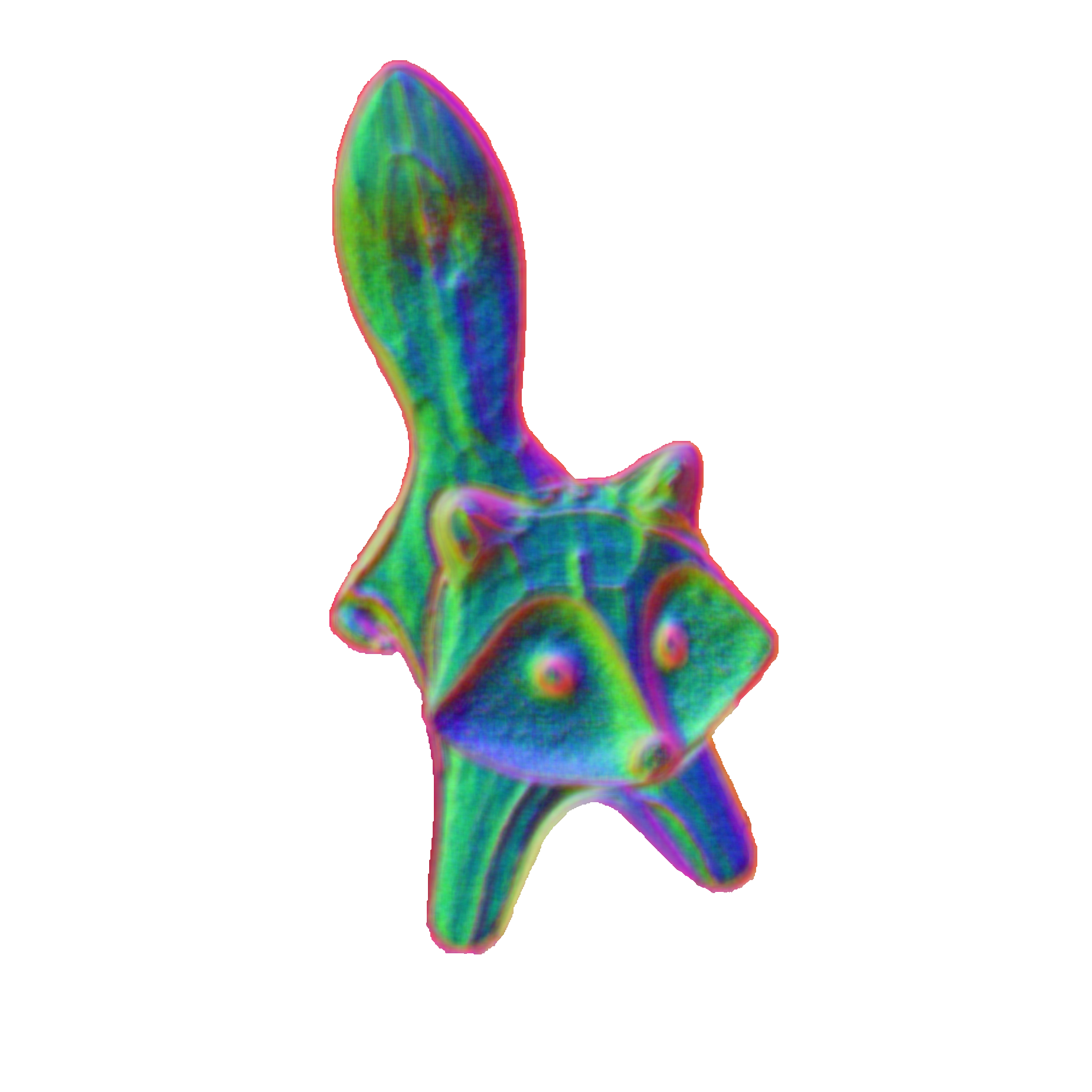}
            \includegraphics[width=0.24\linewidth]{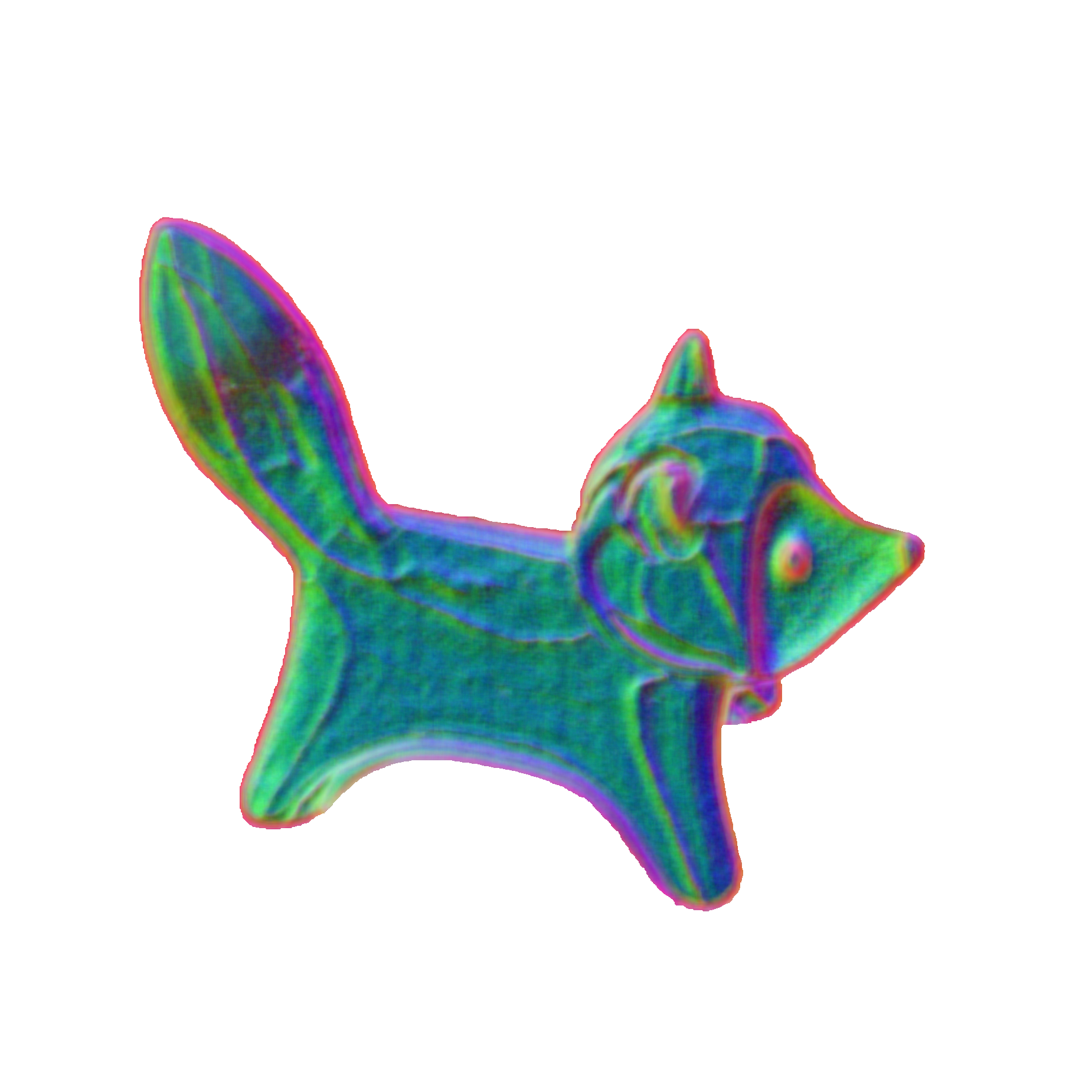}
            \includegraphics[width=0.24\linewidth]{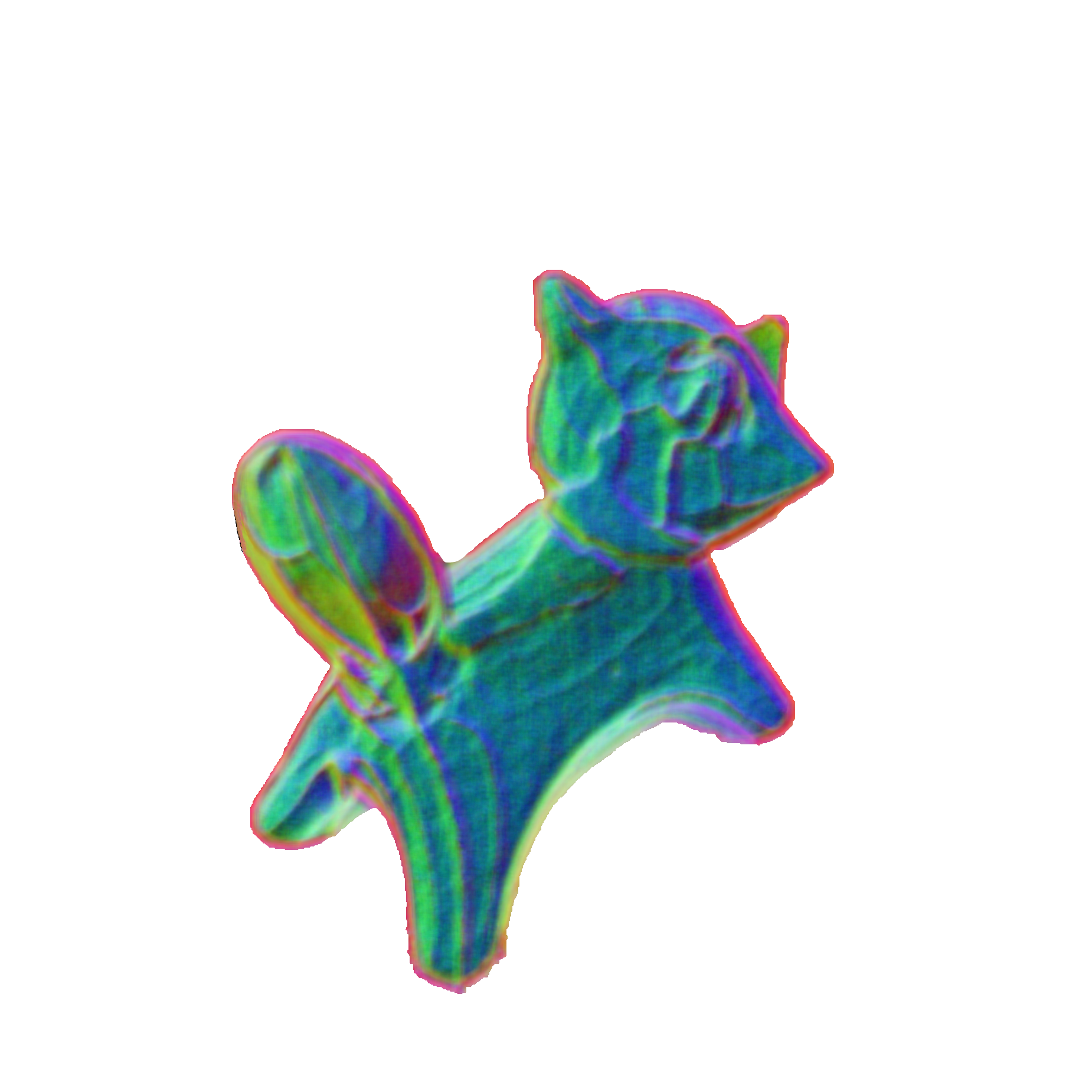}
            \includegraphics[width=0.24\linewidth]{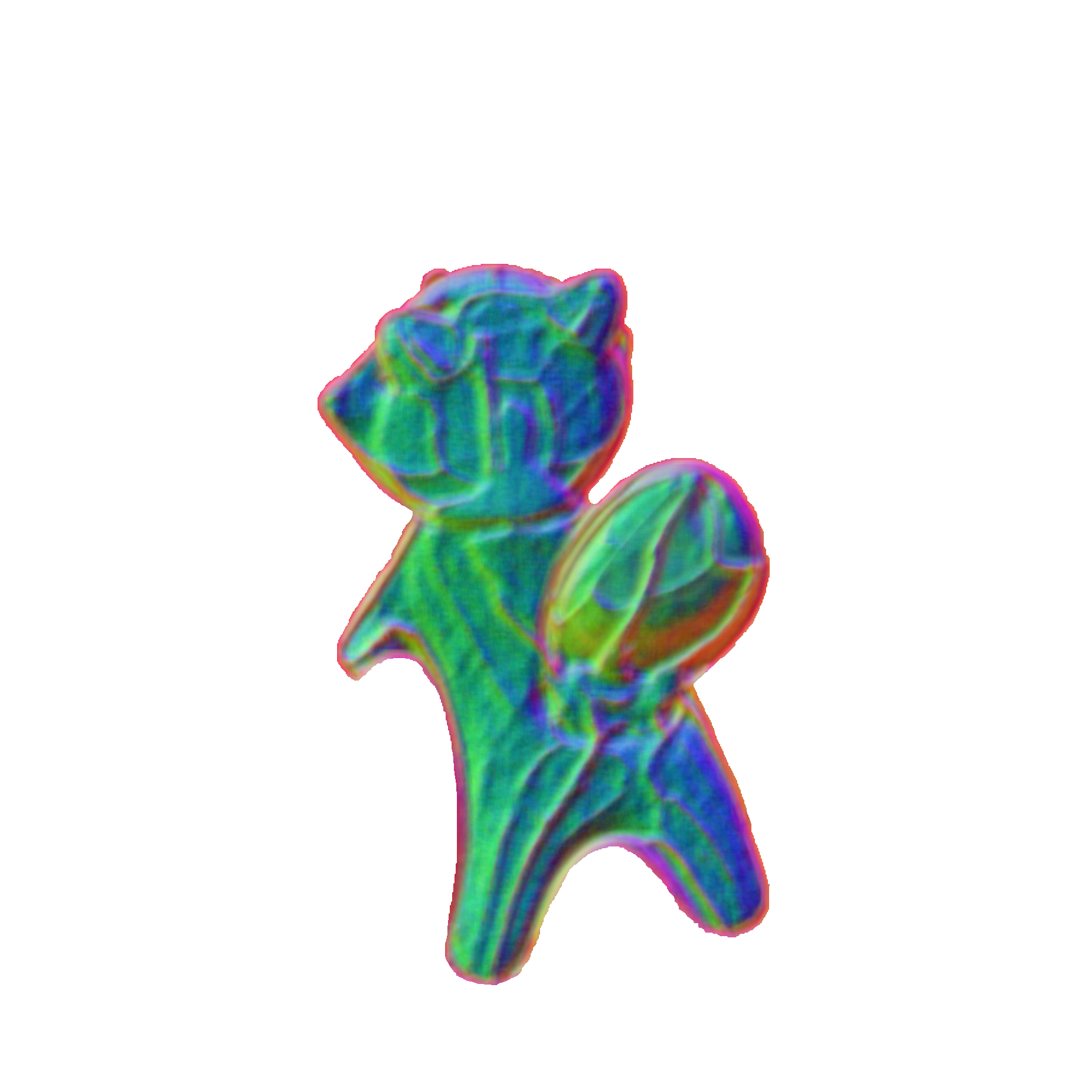}
      \end{minipage}
      
      \begin{minipage}{0.05in}	
	    	   \centering
        		\rotatebox{90}{\small Ours}
      \end{minipage}	
      \hspace{0.01in}	
      \begin{minipage}{3.1in}
         % \begin{subfigure}[b]{\linewidth}
            \centering
            \includegraphics[width=0.24\linewidth]{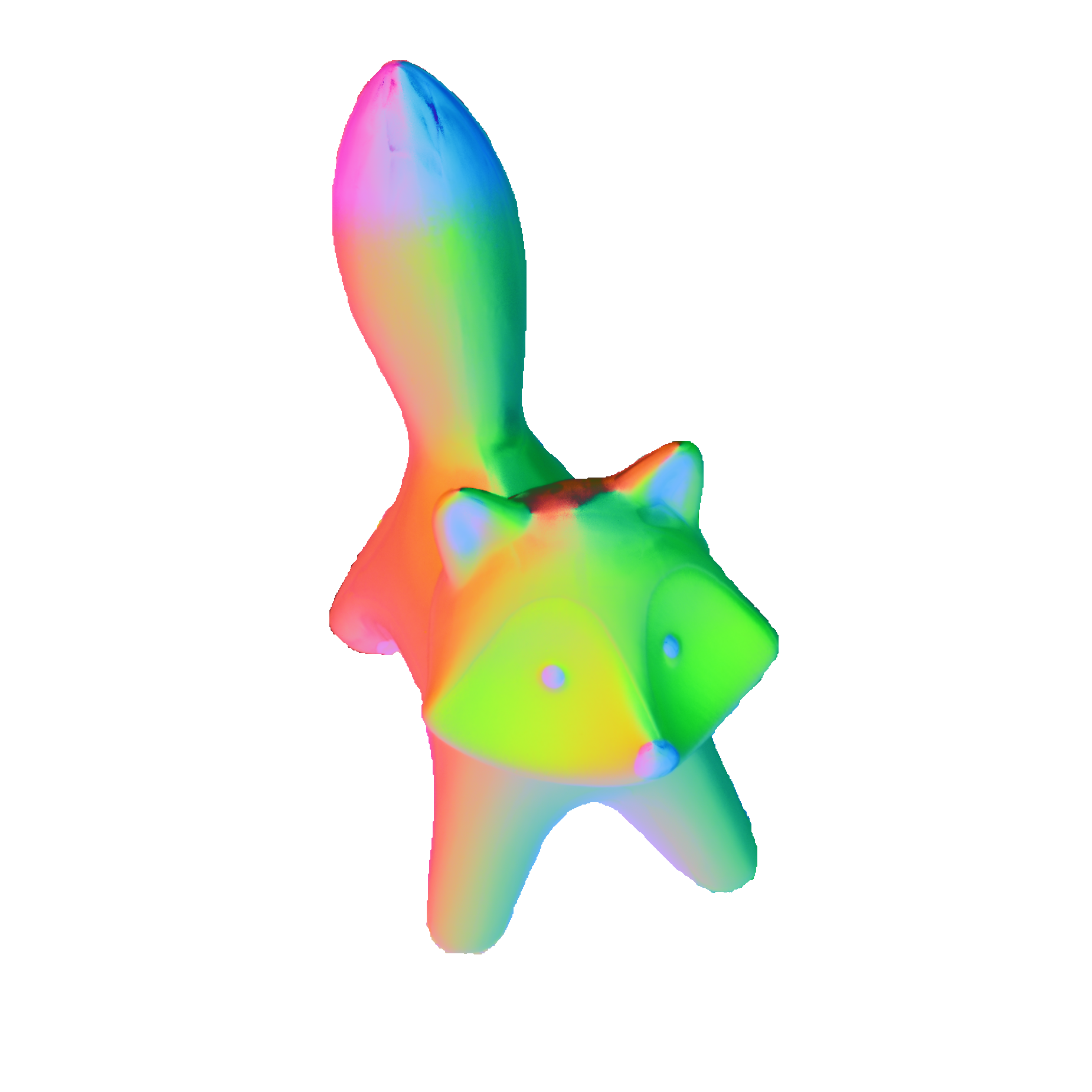}
            \includegraphics[width=0.24\linewidth]{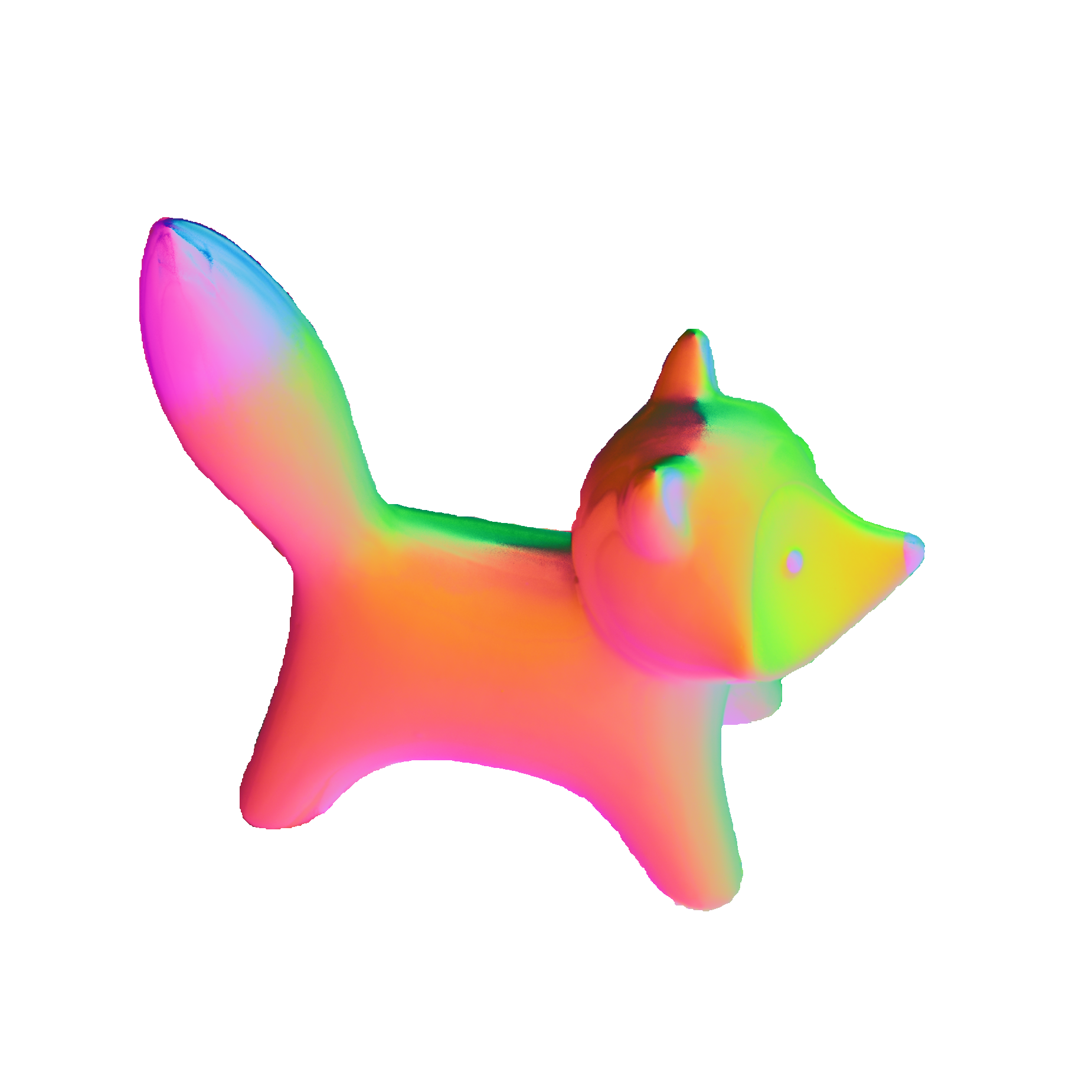}
            \includegraphics[width=0.24\linewidth]{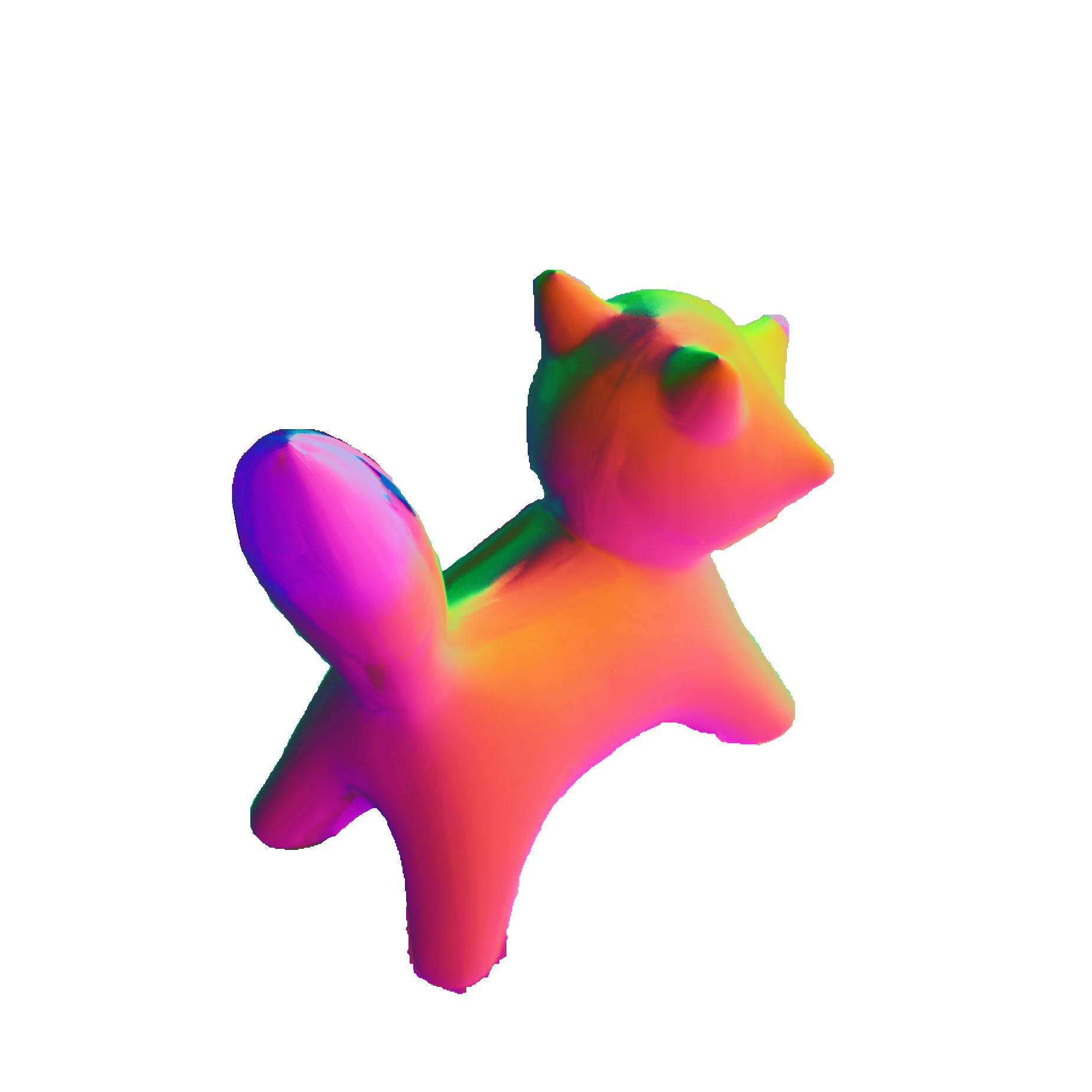}
            \includegraphics[width=0.24\linewidth]{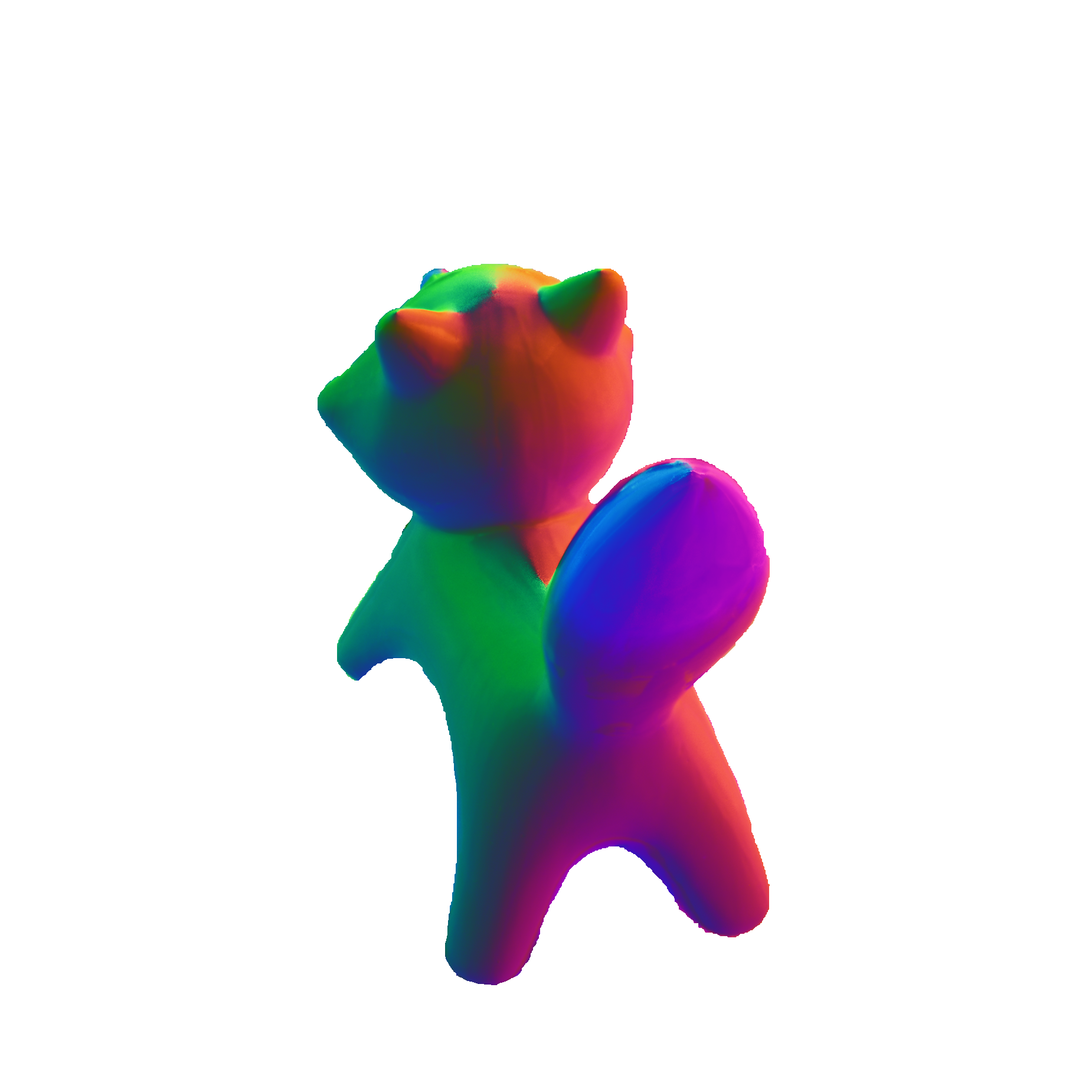}
       \end{minipage}

   \caption{Feature comparisons at different views. The top row are photographs under a full-on lighting pattern. The second row visualizes the features learned by L2-Net~\cite{tian2017l2} from the above photographs. Our features are visualized in the last row, which are more consistent across multiple views. All high-dimensional features are projected to 3D for visualization.}
   \label{fig:ours_l2net}
\end{figure}

We visualize our transformed per-pixel features  at different views in~\figref{fig:ours_l2net}. For comparison, we also visualize the learned local patch descriptors of L2-Net~\cite{tian2017l2}, computed from photographs taken with a full-on lighting pattern to physically reduce the view/lighting variance of the appearance. We directly test their network pre-trained on the Brown dataset~\cite{brown2010discriminative}. Note that our features are more view consistent compared with L2-Net, resulting in more correspondence matches as shown in the next subsection.

\subsection{Comparisons}

\begin{figure}         

         \begin{minipage}{0.3\linewidth}
            \centering
            {\small Single-view~\cite{ikehata2014photometric}}
         \end{minipage}		
         \begin{minipage}{0.3\linewidth}
            \centering
            {\small Multi-view~\cite{li2020multi}}
         \end{minipage}		
         \begin{minipage}{0.3\linewidth}
            \centering
            {\small Ours}
         \end{minipage}		
         \begin{minipage}{0.056\linewidth}
         	\hspace{0in}
         \end{minipage}	   
   
      % \hfill
      \includegraphics[width=0.3\linewidth]{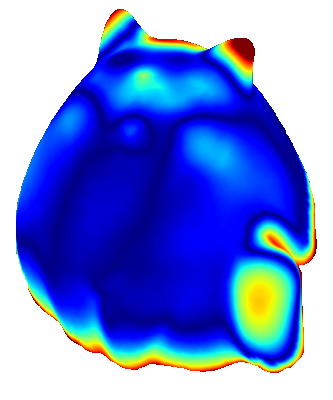}
      \includegraphics[width=0.3\linewidth]{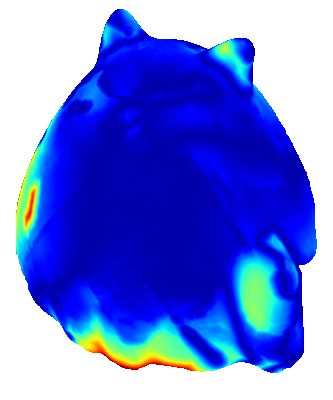}
      \includegraphics[width=0.3\linewidth]{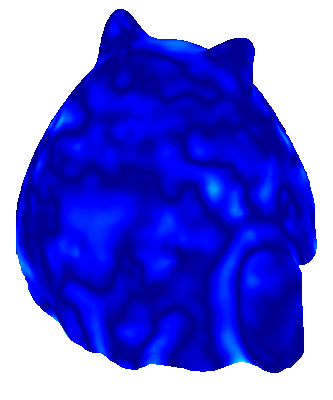}
      \begin{minipage}{0.056\linewidth}
      	\vspace{-1.1in}
      	\includegraphics[width=\linewidth]{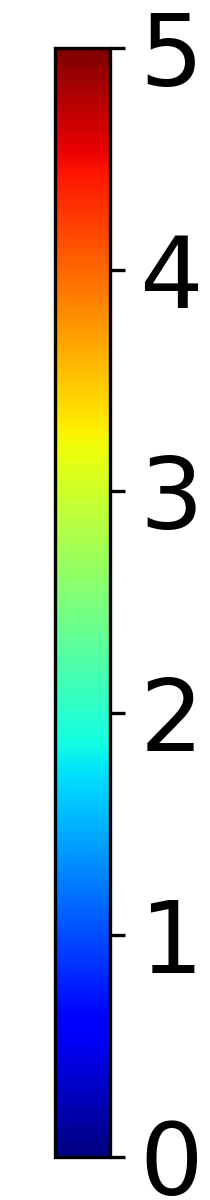}
      \end{minipage}	   

         \begin{minipage}{0.3\linewidth}
            \centering
            {\small 12.8/11.2}
         \end{minipage}		
         \begin{minipage}{0.3\linewidth}
            \centering
            {\small 28.3/65.0}
         \end{minipage}		
         \begin{minipage}{0.3\linewidth}
            \centering
            {\small \textbf{38.5/83.6}}
         \end{minipage}		
         \begin{minipage}{0.056\linewidth}
         	\hspace{0in}
         \end{minipage}	   

   \caption{Comparison between single-~\cite{ikehata2014photometric} / multi-view~\cite{li2020multi} photometric stereo and our framework with a lightstage. We show color-coded geometric reconstruction errors with respect to a ground-truth model. The unit of the scale bar is mm. Quantitative errors in accuracy / completeness are reported at the bottom.}
   \label{fig:cmp_spf_mpf}
\end{figure}

\begin{figure}             
         \centering

         \begin{minipage}{0.32\linewidth}
            \centering
            {\small Single-view~\cite{chen2018ps}}
         \end{minipage}		
         \begin{minipage}{0.32\linewidth}
            \centering
            {\small Multi-view~\cite{li2020multi}}
         \end{minipage}		
         \begin{minipage}{0.32\linewidth}
            \centering
            {\small Ours}
         \end{minipage}		
       
         % \hfill
            \centering
            % \rotatebox{90}{\small BEAR}
            % \hspace{-2in}
            \includegraphics[width=0.32\linewidth]{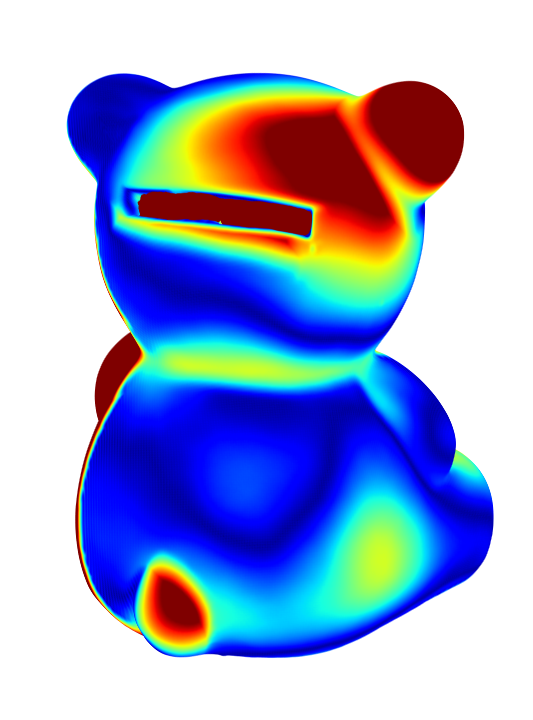}
            \includegraphics[width=0.32\linewidth]{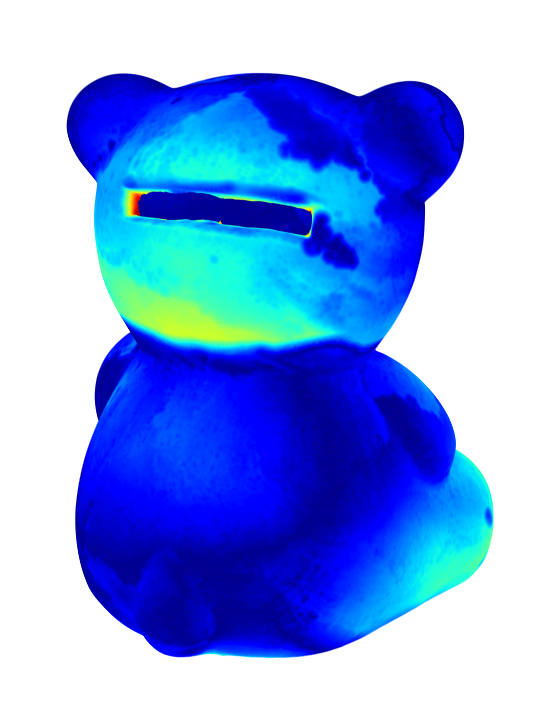}
            \includegraphics[width=0.32\linewidth]{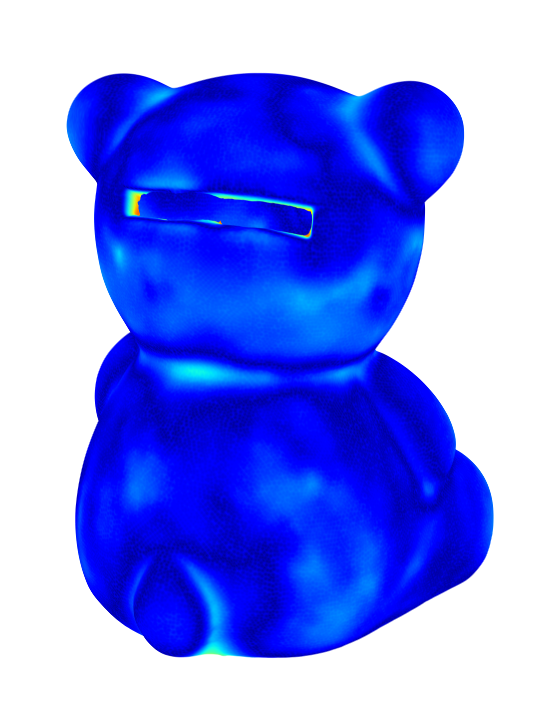}

         \begin{minipage}{0.32\linewidth}
            \centering
            {\small 24.8/28.4}
         \end{minipage}		
         \begin{minipage}{0.32\linewidth}
            \centering
            {\small 49.7/79.6}
         \end{minipage}		
         \begin{minipage}{0.32\linewidth}
            \centering
            {\small \textbf{53.5/83.9}}
         \end{minipage}		
         
      \caption{Comparison between photometric stereo and our framework adapted to the input from DiLiGenT-MV. We show color-coded geometric reconstruction errors.  Quantitative errors in accuracy / completeness are reported at the bottom.}
      \label{fig:cmp_diligent}
\end{figure}   
\begin{table}[htb]
   \begin{center}
    \caption{Quantitative evaluation of the reconstruction quality of different methods (in~\figref{fig:cmp_mvs}), in accuracy / completeness (\%) at a 1mm threshold.}
    \label{table:error}
    \setlength{\tabcolsep}{0.5mm}{
       \begin{tabular}{crrrrr}
          \hline         
          \makecell[c]{}  & \makecell[c]{Ours}  &  \makecell[c]{\cite{guo2020cascade}} &  \makecell[c]{\cite{tian2017l2}}&  \makecell[c]{\cite{Kang:2019:JOINT}}&  \makecell[c]{\cite{schoenberger2016sfm,schoenberger2016mvs}}\\
          \hline
          \small Fox & 
         \textbf{53.1}/\textbf{71.9} & 41.1/34.2 & 29.4/20.0 & 40.2/26.9 & 39.9/21.3 \\
          \small Cat & 
          \textbf{38.5}/\textbf{83.6} & 33.5/23.8 & 29.7/08.8  & 28.1/13.1 & 38.4/05.7 \\
          \small Rooster &
           65.3/\textbf{79.8} & 66.5/60.6 & 55.8/53.8  & 56.3/19.8 & \textbf{70.6}/23.1  \\
          \small Cup &
           44.6/\textbf{98.2} & 44.6/75.4  & 34.1/39.2 & 41.5/34.4 & \textbf{51.4}/29.5  \\
          \hline
       \end{tabular}
    }
    \end{center}
 \end{table}
 
\begin{figure*}[htb]
      \begin{minipage}{\textwidth}
         \begin{minipage}{0.05in}
               \hspace{-2in}        	
         \end{minipage}	
         \begin{minipage}{6.9in}
            \centering
            \begin{minipage}{0.155\linewidth}
               \centering
               {\small Photograph}
            \end{minipage}		
            \begin{minipage}{0.155\linewidth}
               \centering
               {\small Ours}
            \end{minipage}		
            \begin{minipage}{0.155\linewidth}
               \centering
               {\small CasMVSNet~\cite{guo2020cascade}}
            \end{minipage}		
            \begin{minipage}{0.155\linewidth}
               \centering
               {\small L2-Net~\cite{tian2017l2}}
            \end{minipage}		
            \begin{minipage}{0.155\linewidth}
               \centering
               {\small Albedos/Normals\cite{Kang:2019:JOINT}}
            \end{minipage}		
            \begin{minipage}{0.155\linewidth}
               \centering
               {\small COLMAP~\cite{schoenberger2016sfm,schoenberger2016mvs}}
            \end{minipage}		
      \end{minipage}       

       \begin{minipage}{0.05in}	
        		\rotatebox{90}{\small Fox}
		      \hspace{-2in}
      \end{minipage}	
      \begin{minipage}{6.9in}
            \centering
            \includegraphics[width=0.155\linewidth]{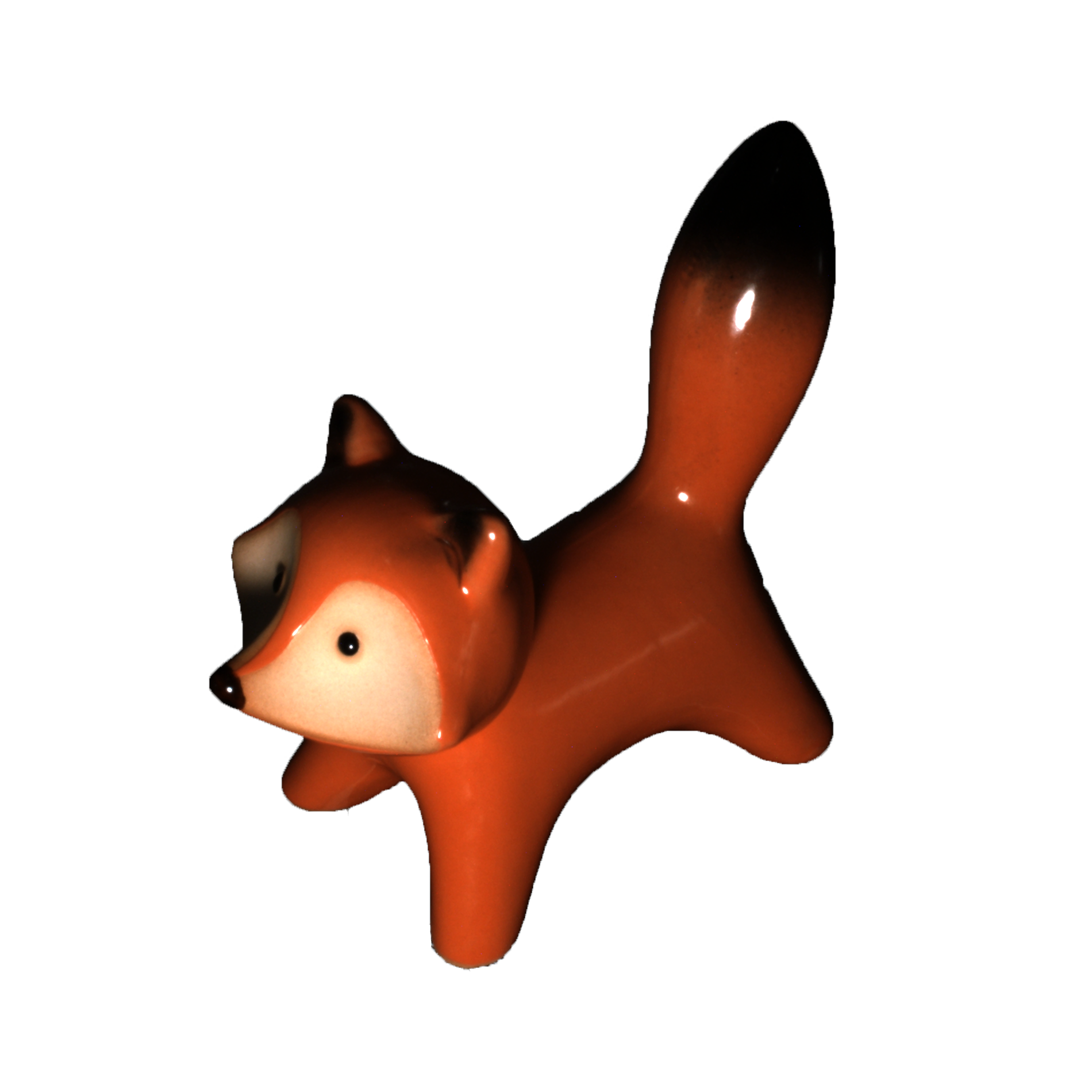}
            \includegraphics[width=0.155\linewidth]{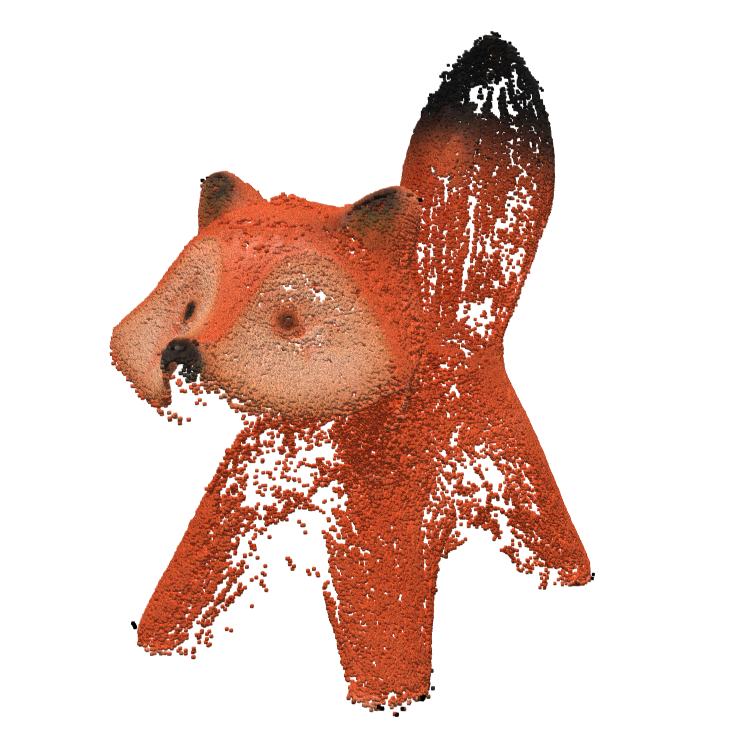}
            \includegraphics[width=0.155\linewidth]{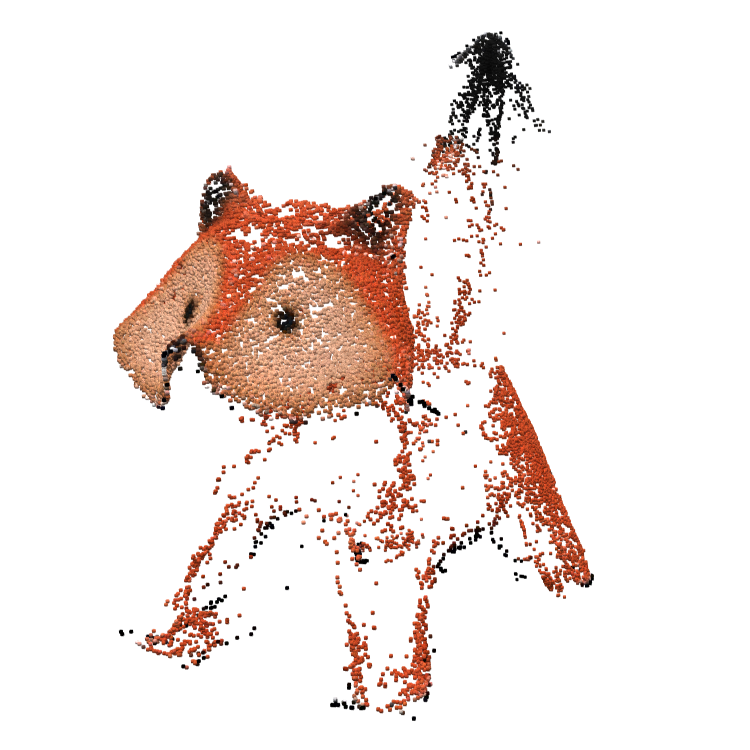}
            \includegraphics[width=0.155\linewidth]{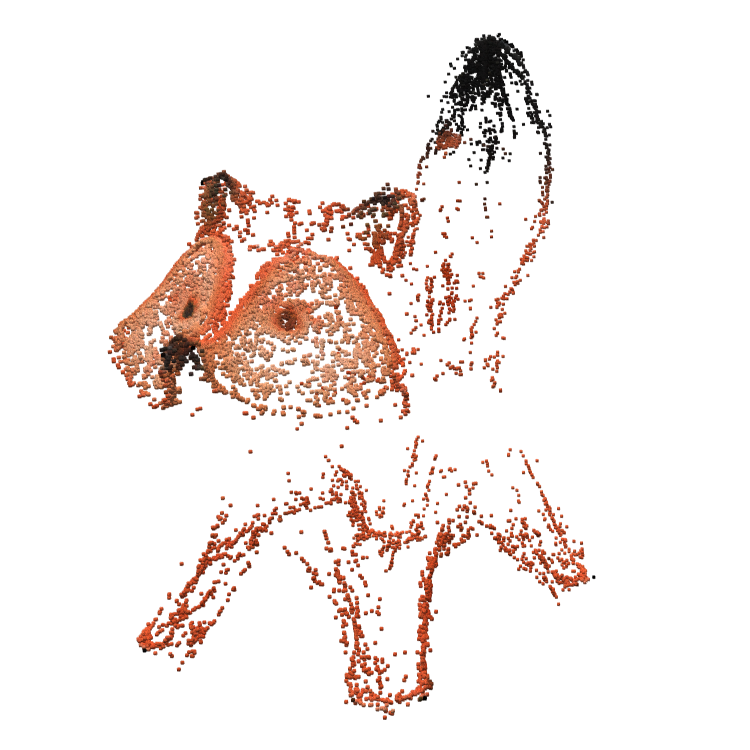}
            \includegraphics[width=0.155\linewidth]{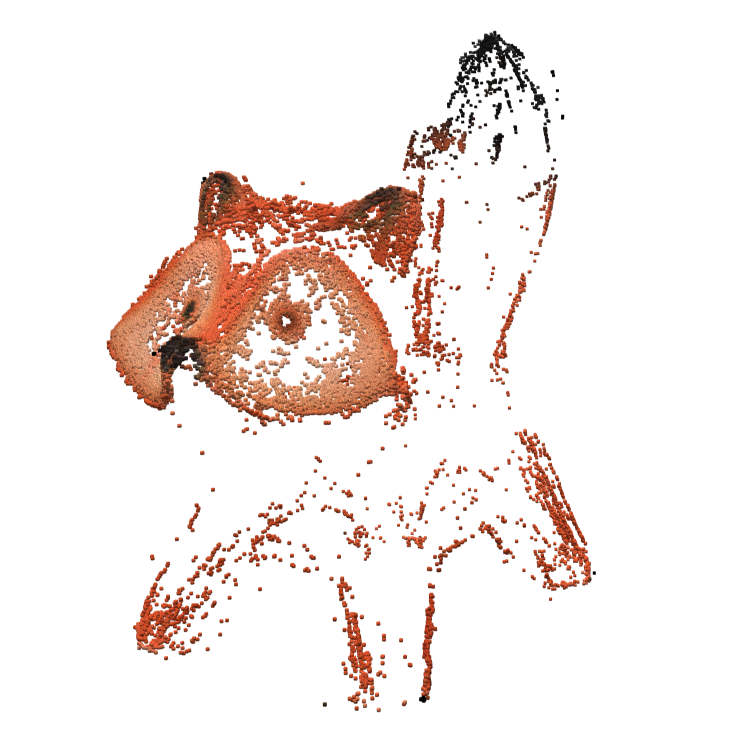}
            \includegraphics[width=0.155\linewidth]{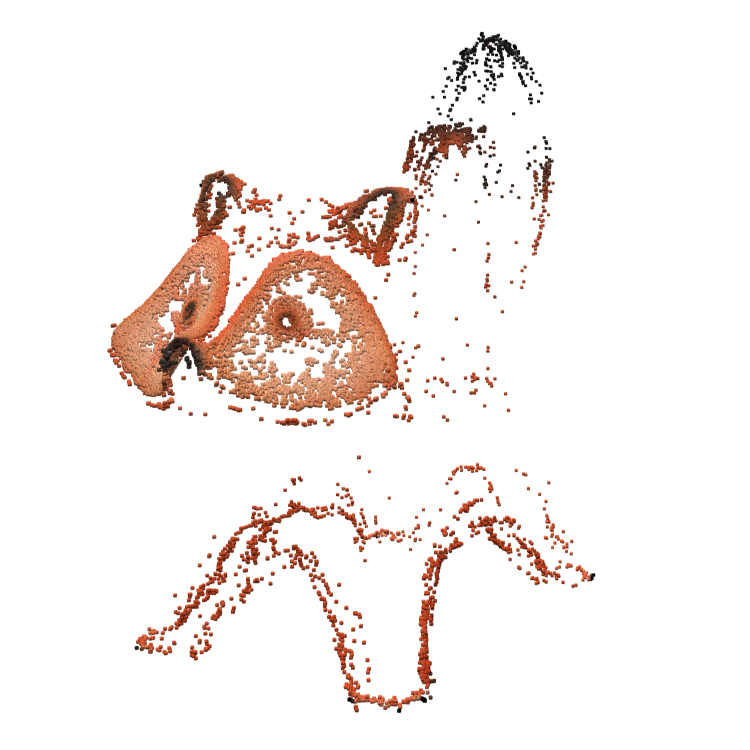}
      \end{minipage}

      \begin{minipage}{0.05in}	
	    	   \centering
        		\rotatebox{90}{\small Cat}
		      \hspace{-2in}
      \end{minipage}	
      \begin{minipage}{6.9in}
            \centering
            \includegraphics[width=0.155\linewidth]{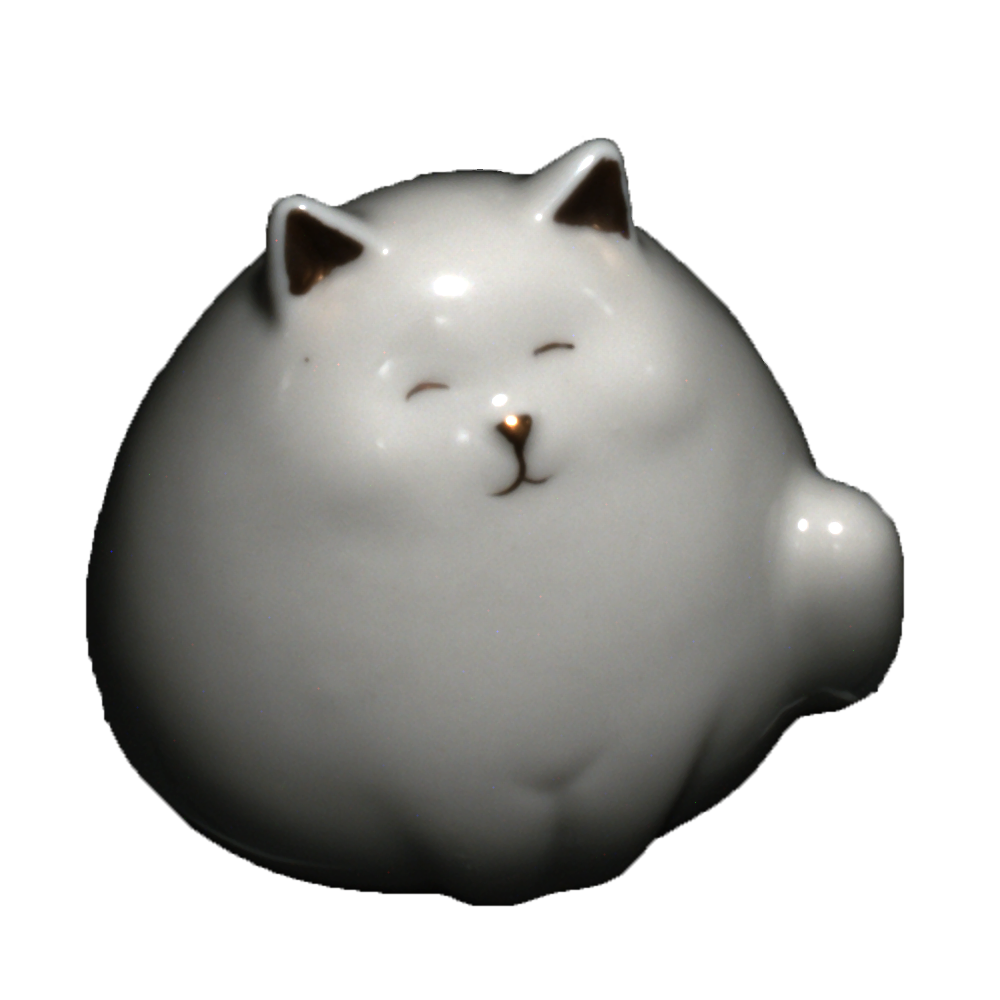}
            \includegraphics[width=0.155\linewidth]{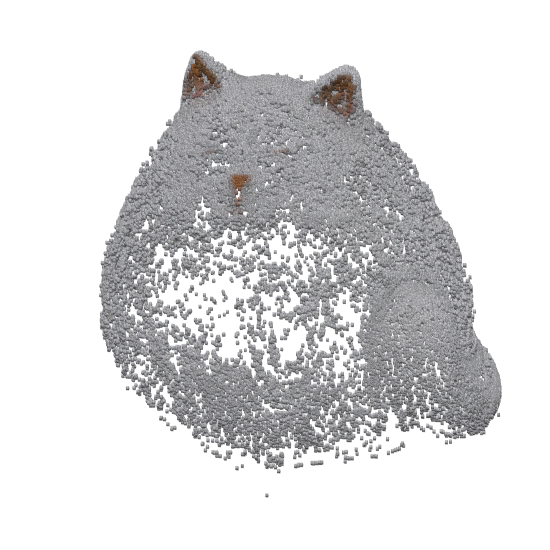}
            \includegraphics[width=0.155\linewidth]{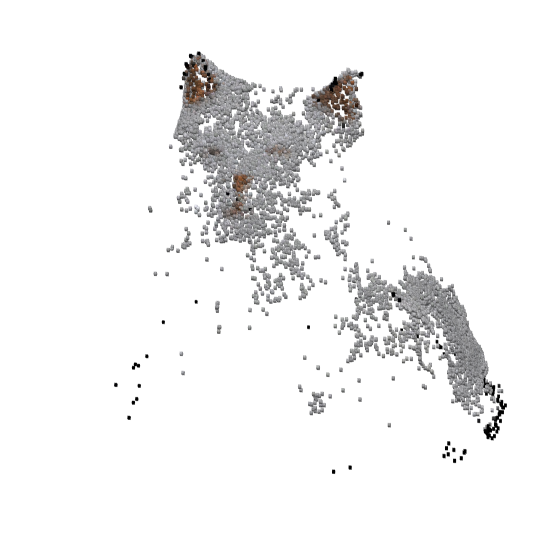}
            \includegraphics[width=0.155\linewidth]{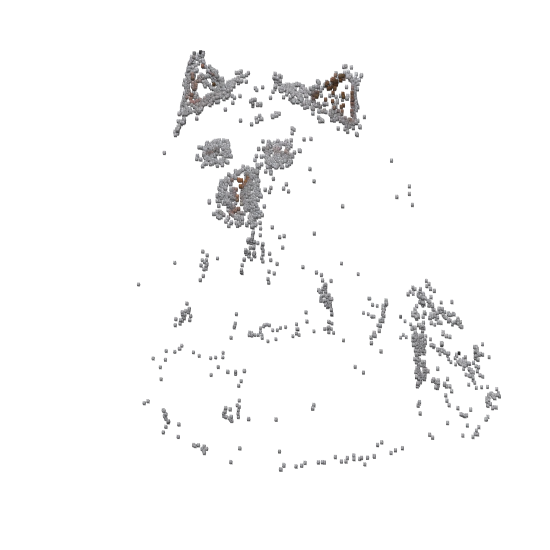}
            \includegraphics[width=0.155\linewidth]{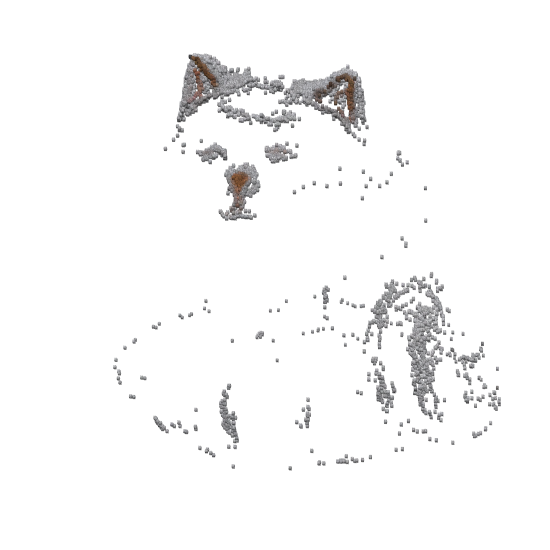}
            \includegraphics[width=0.155\linewidth]{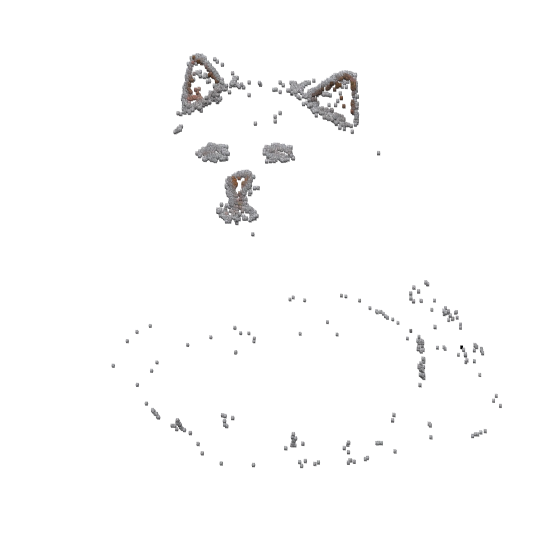}
      \end{minipage}
      
      \begin{minipage}{0.05in}	
	    	   \centering
        		\rotatebox{90}{\small Rooster}
		      \hspace{-2in}
      \end{minipage}	
      \begin{minipage}{6.9in}
            \centering
            \includegraphics[width=0.155\linewidth]{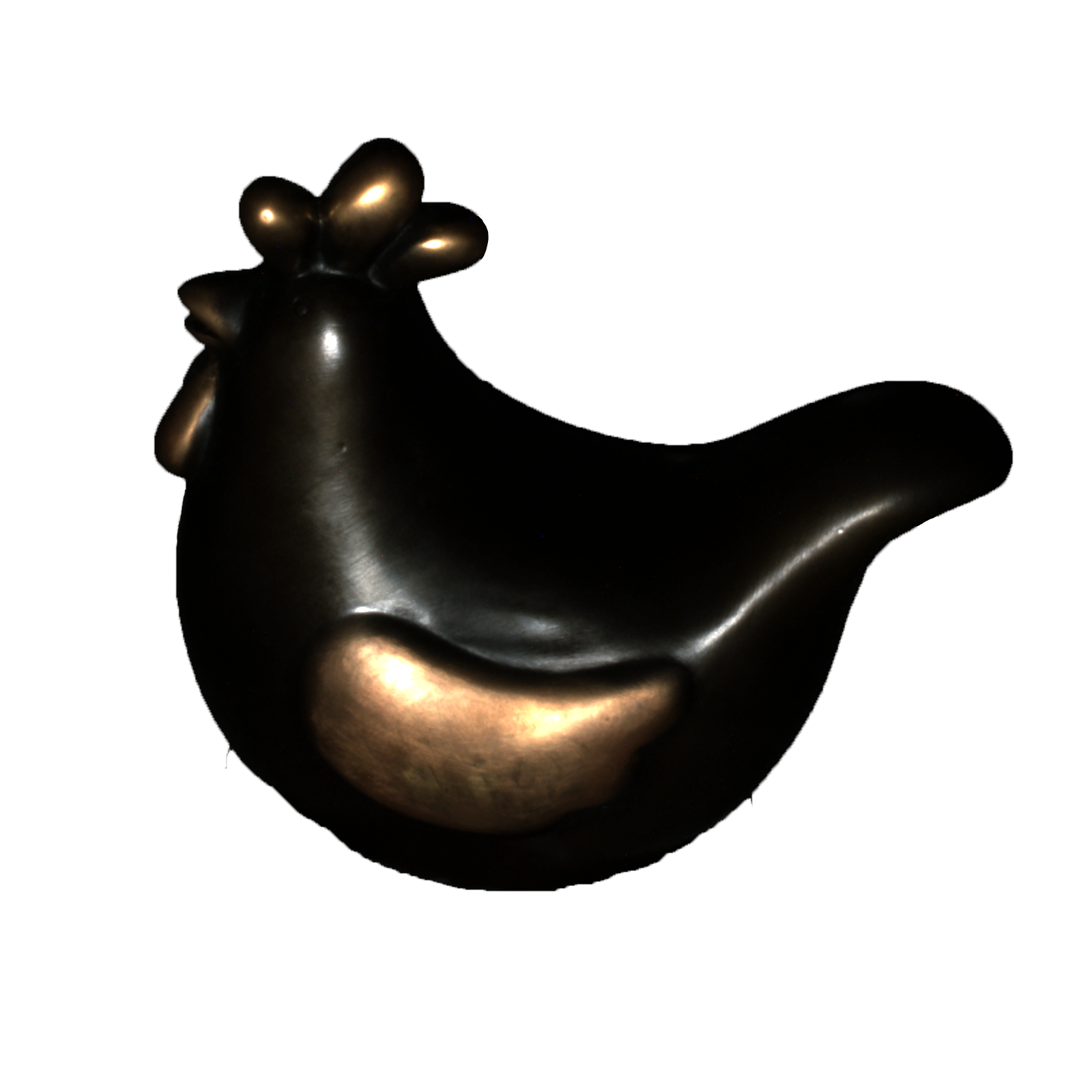}
            \includegraphics[width=0.155\linewidth]{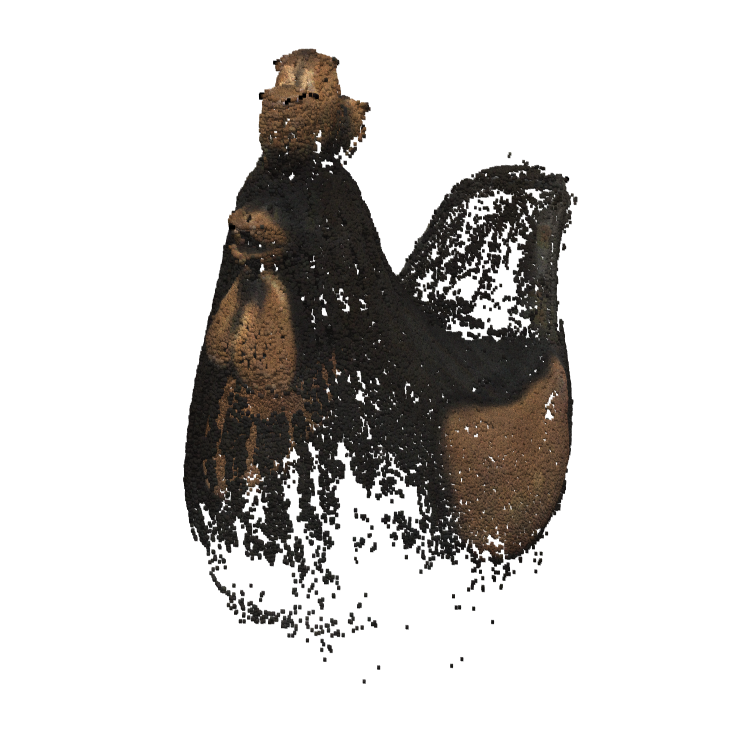}
            \includegraphics[width=0.155\linewidth]{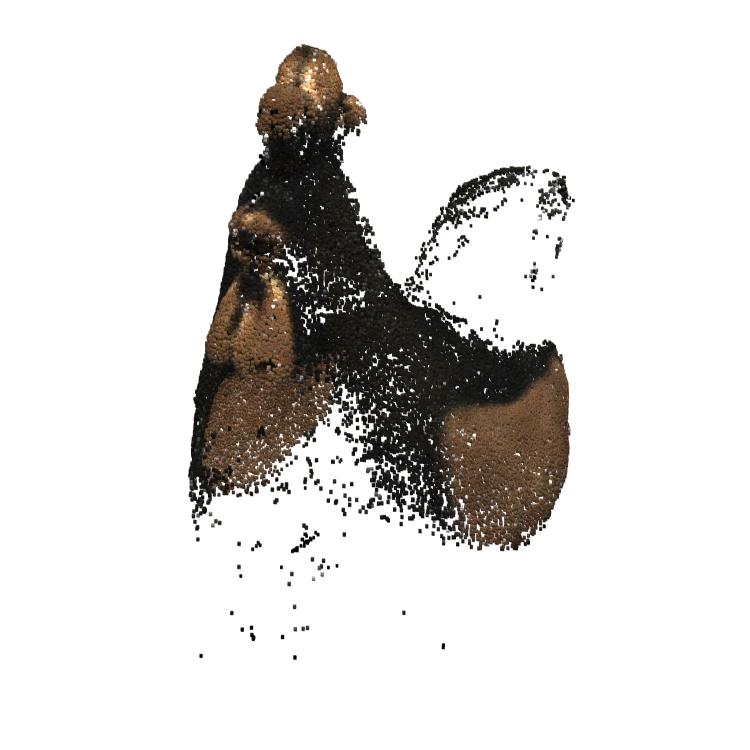}
            \includegraphics[width=0.155\linewidth]{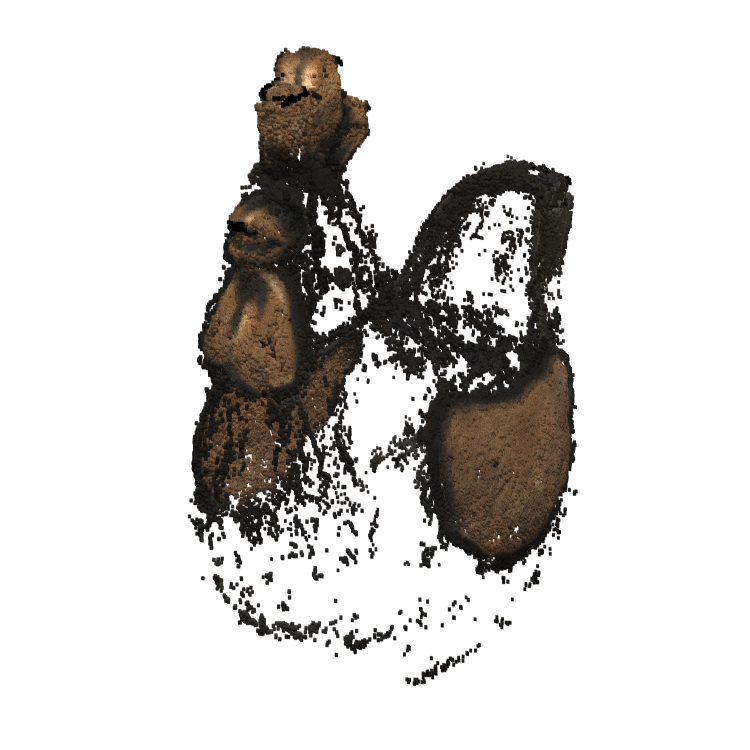}
            \includegraphics[width=0.155\linewidth]{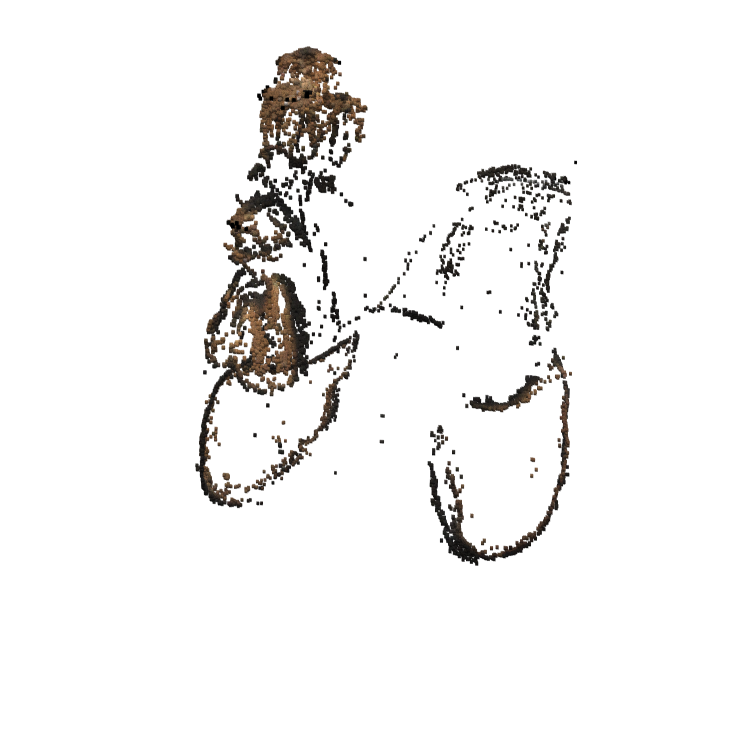}
            \includegraphics[width=0.155\linewidth]{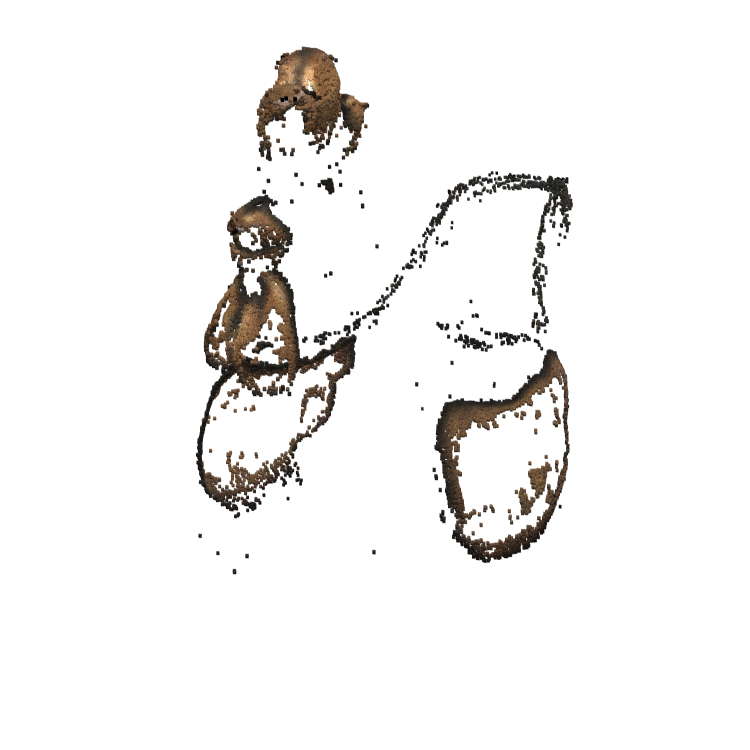}
      \end{minipage}      
      
      \begin{minipage}{0.05in}	
	    	   \centering
        		\rotatebox{90}{\small Cup}
		      \hspace{-2in}
      \end{minipage}	
      \begin{minipage}{6.9in}
            \centering
            \includegraphics[width=0.155\linewidth]{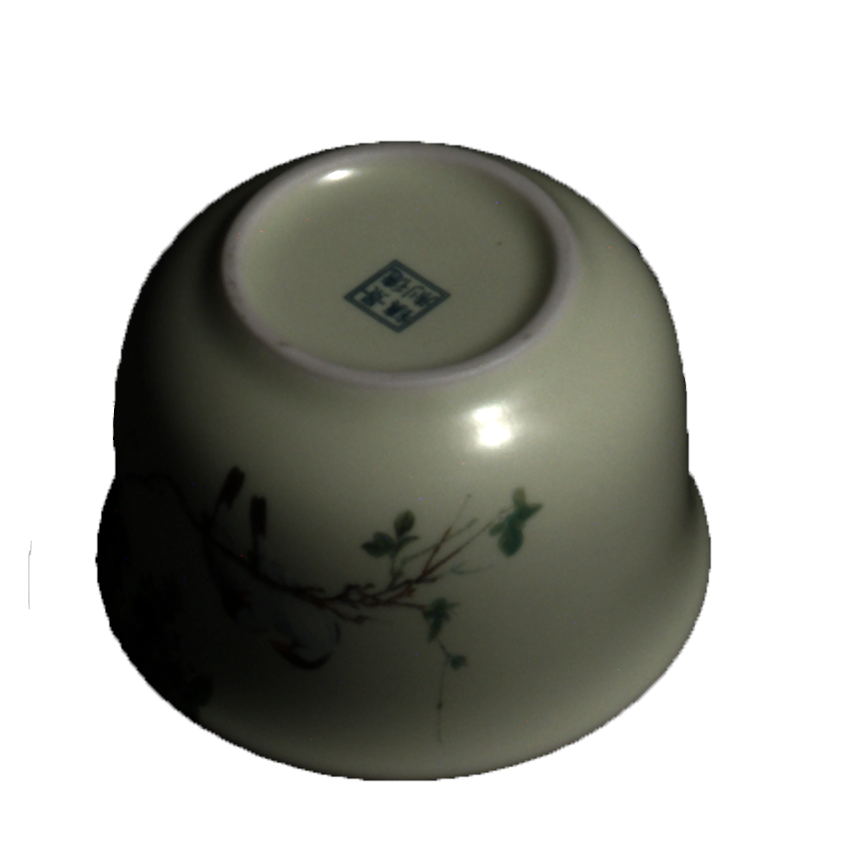}
            \includegraphics[width=0.155\linewidth]{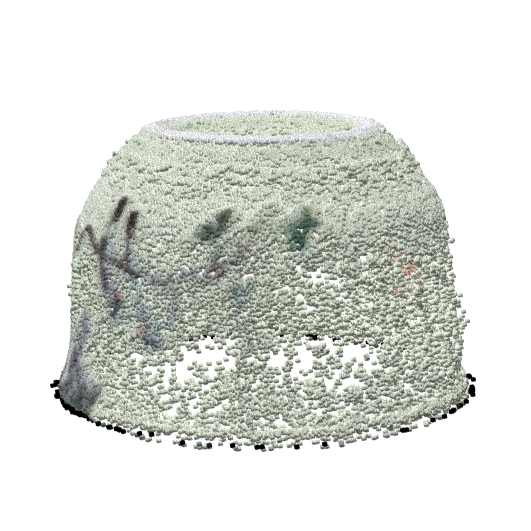}
            \includegraphics[width=0.155\linewidth]{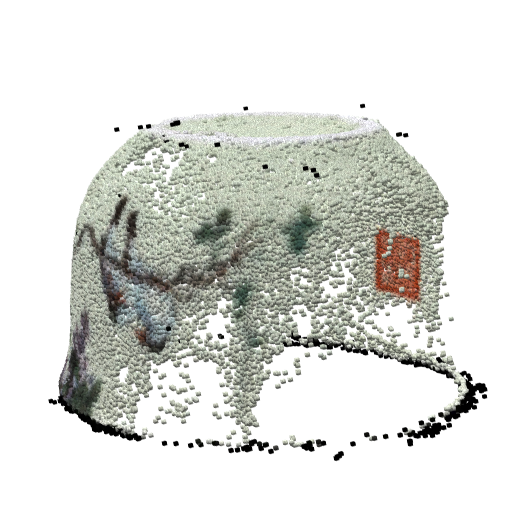}
            \includegraphics[width=0.155\linewidth]{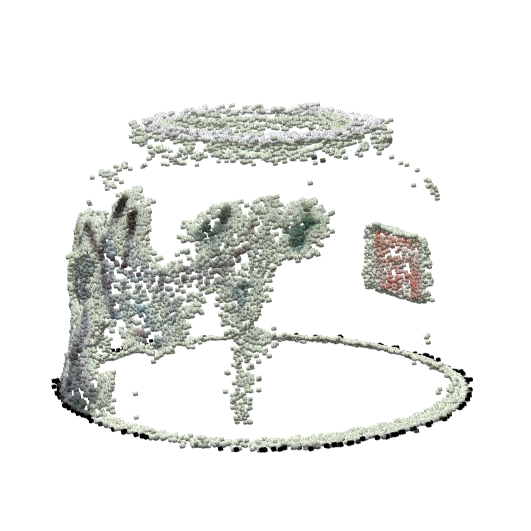}
            \includegraphics[width=0.155\linewidth]{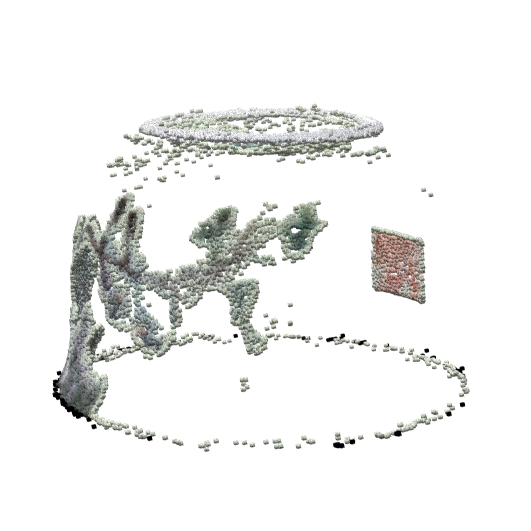}
            \includegraphics[width=0.155\linewidth]{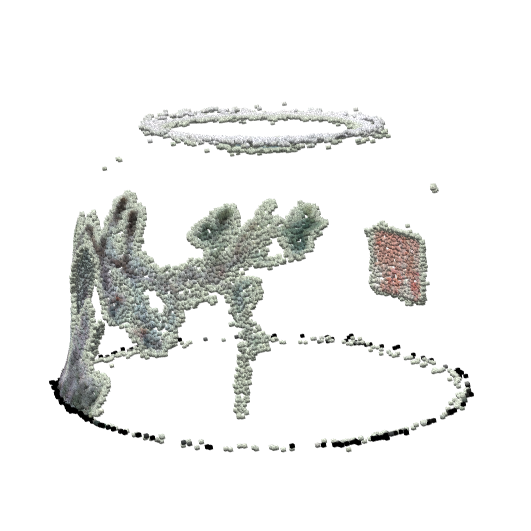}
      \end{minipage}

   \end{minipage}
   \caption{Comparisons with multi-view stereo methods. The first column are photographs of the physical objects. The second column to the last are geometric reconstruction results using our learned features, CasMVSNet~\cite{li2020multi}, learned local features using L2-Net~\cite{tian2017l2},  combined results from diffuse albedo / normal maps according to~\cite{Kang:2019:JOINT}, and COLMAP~\cite{schoenberger2016sfm,schoenberger2016mvs} on images taken with a full-on lighting pattern. Corresponding quantitative errors are reported in Tab.~\ref{table:error}.}
   \label{fig:cmp_mvs}
\end{figure*}

\textbf{Multi-View Stereo.} In~\figref{fig:cmp_mvs}, we compare the reconstruction results of a number of objects captured with the lightstage, using our framework against four alternative methods in multi-view stereo. CasMVSNet~\cite{guo2020cascade} is a state-of-the-art learning-based multi-view stereo technique. L2-Net~\cite{tian2017l2} learns local features for 3D reconstruction, while differentiable reflectance capture~\cite{Kang:2019:JOINT} employs diffuse albedos and normals as invariant geometric features. For COLMAP~\cite{schoenberger2016sfm,schoenberger2016mvs}, we feed multi-view photographs taken with a full-on lighting pattern. In all cases,  existing work mainly generates 3D points around regions with high spatial frequencies. For other regions, reliable correspondences cannot be established due to the lack of discriminative input information. Our method produces more complete shapes, as we fully and automatically exploit the photometric information for computing distinctive and view-invariant features (visualized in~\figref{fig:ours_l2net}). Please refer to Tab.~\ref{table:error} for quantitative errors measured in accuracy / completeness (\%) at a 1mm threshold. 

\textbf{Photometric Stereo.} We first compare our network adapted to the data from the lightstage, against state-of-the-art single-~\cite{ikehata2014photometric} / multi-view~\cite{li2020multi} photometric stereo in~\figref{fig:cmp_spf_mpf}. One hundred photographs with different illumination conditions are fed to both methods as input. To perform quantitative error analysis on the reconstructed geometry, we acquire a ground-truth shape, using a structured-lighting-based handheld industrial scanner~\cite{EinScanP93:online}, with an accuracy of $0.05$mm. In comparison with related work, our reconstruction result has a lower geometric error, due to a wider coverage of reliable features correspondences (cf.~\figref{fig:cmp_mvs}).

Next, we compare our network generalized to the input of DiLiGenT-MV (\sec{sec:details}) against photometric stereo on the Bear object. Our result compares favorably with~\cite{chen2018ps} and~\cite{li2020multi}. For~\cite{chen2018ps}, the integration of normals accumulates errors in depth map generation, which are challenging to correct at a single view. For~\cite{li2020multi}, their quality is sensitive to the accuracy of initial points, and its depth propagation is affected by the curvature. This experiment demonstrates our generalization ability to different input conditions / acquisition setups.
   
\begin{figure}
      \centering
      \includegraphics[width=\linewidth]{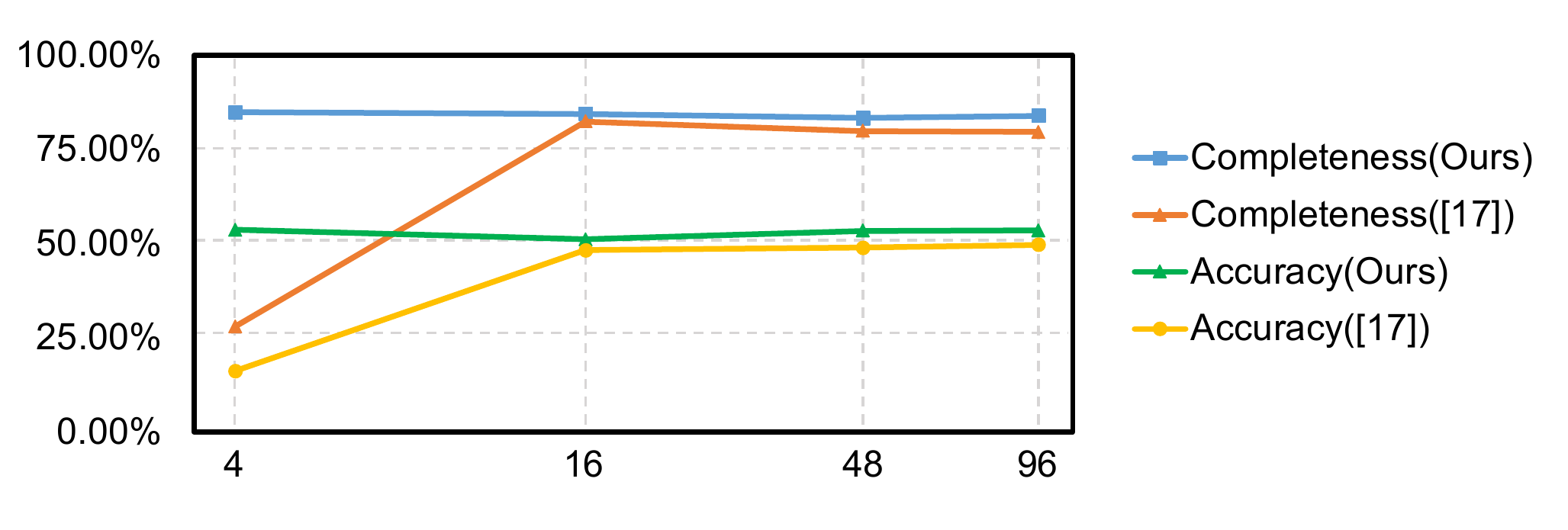}
   \caption{Comparison of reconstruction accuracy / completeness of Bear with multi-view photometric stereo~\cite{li2020multi} using different number of input lighting conditions, shown on the x axis.}
   \label{fig:curve}
\end{figure}

\begin{figure}
   \begin{minipage}{\linewidth}
      \centering
      \includegraphics[height=1.12in]{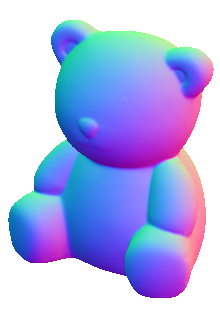}
      \includegraphics[height=1.12in]{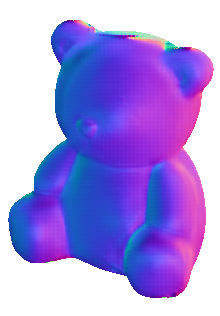}
      \includegraphics[height=1.12in]{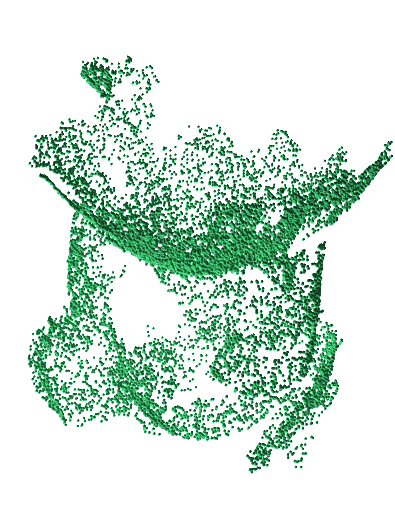}
      \includegraphics[height=1.12in]{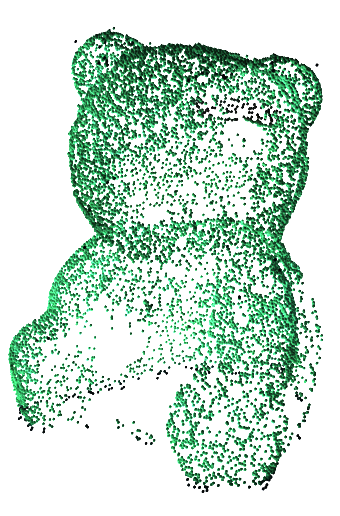}
   \end{minipage}
   \caption{Comparison with photometric stereo with only 4 input lighting conditions. From the left to right, the ground-truth normal map, predicted normals using~\cite{chen2018ps}, geometric reconstruction with~\cite{li2020multi}, and our result.}
   \label{fig:4lights}
\end{figure}

In addition, we compare in~\figref{fig:curve} the reconstruction errors of Bear between our framework and multi-view photometric stereo\cite{li2020multi}, using different number of lighting conditions sampled from DiLiGenT-MV. Our reconstruction quality is consistently higher, and does not degrade as the number of lights decreases. \figref{fig:4lights} further visualize the results. As can be seen from the left two images, the limited lighting variations are not sufficient to predict reliable normals, even using one state-of-the-art learning-based method~\cite{chen2018ps}. On the other hand, ~\cite{li2020multi} struggles to identify the azimuth angle in iso-contours, resulting in shape distortions. In comparison, our reconstruction captures the major geometric features of the original object, demonstrating the ability to efficiently exploit even the highly limited geometric information available in the input.  

\begin{figure}
   \begin{minipage}{\linewidth}
      \centering
      \includegraphics[width=0.32\linewidth]{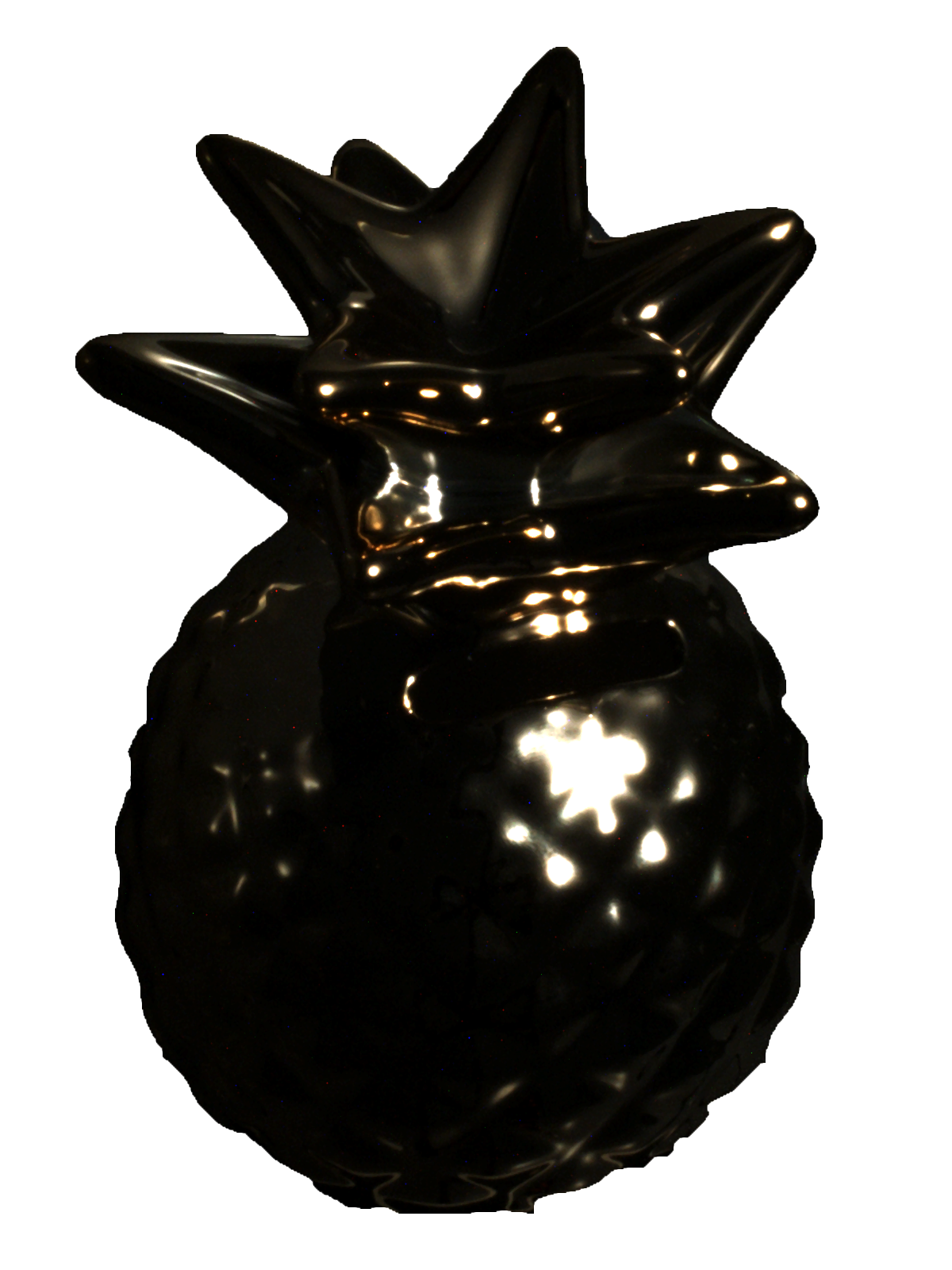}
      \includegraphics[width=0.32\linewidth]{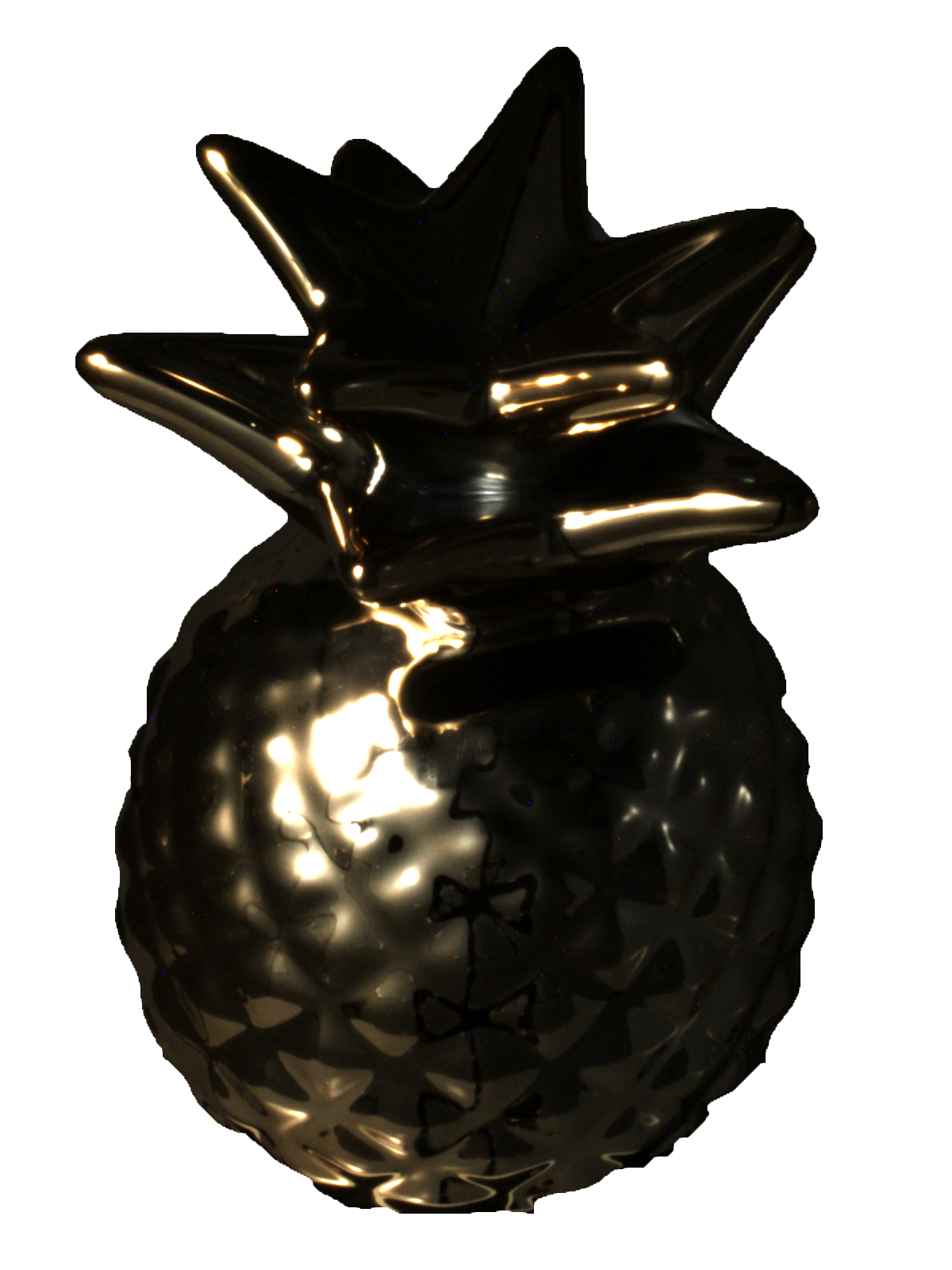}
      \includegraphics[width=0.32\linewidth]{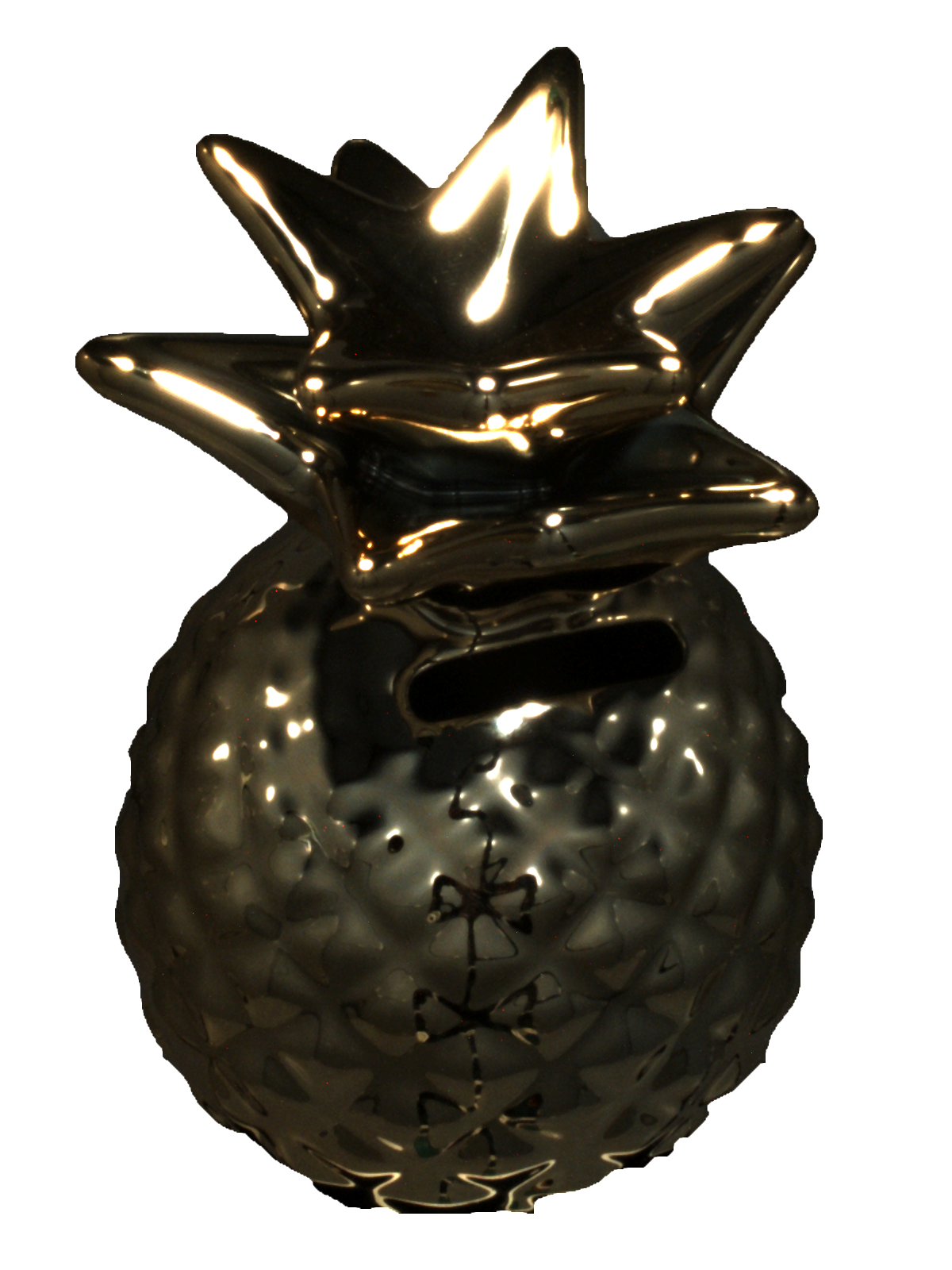}
      \includegraphics[width=0.32\linewidth]{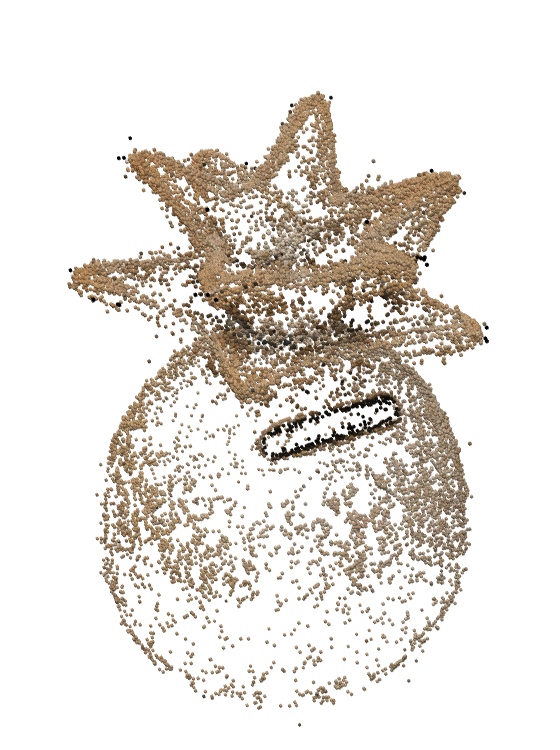}
      \includegraphics[width=0.32\linewidth]{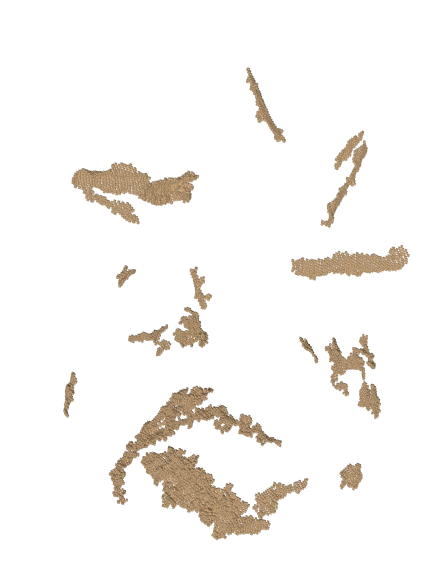}
      \includegraphics[width=0.32\linewidth]{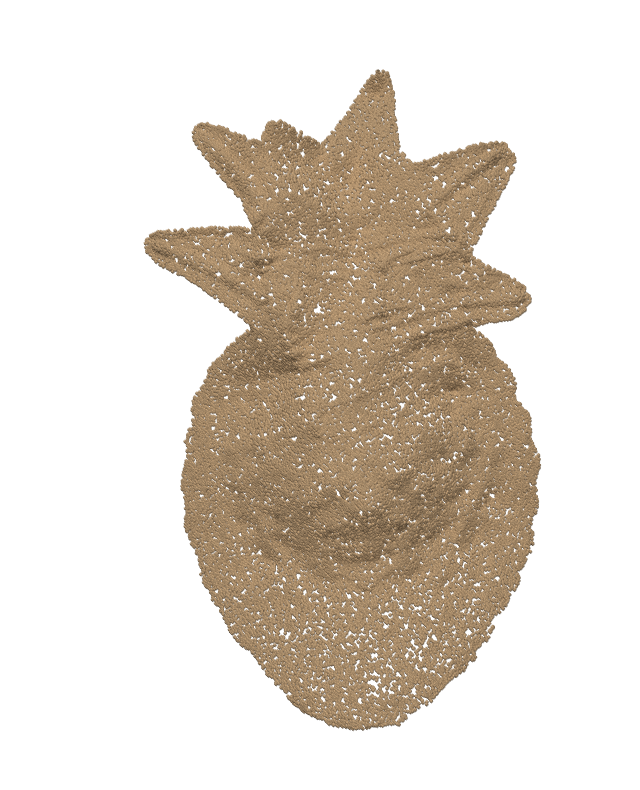}
   \end{minipage}

   \caption{Geometric reconstruction on a challenging object with complex geometry and appearance. The top row shows photographs of the object at the same view but different illumination conditions. From the left to right at the bottom row, our result, the result from a structured-lighting-based 3D scanner~\cite{EinScanP93:online} and single-view photometric stereo~\cite{ikehata2014photometric}.}
   \label{fig:mirrorlike}
\end{figure}

Moreover, we test on an object with highly challenging appearance and shape in~\figref{fig:mirrorlike}. The scanner fails to produce a useful result, due to the complex light transport over the object surfaces that severely distorts the structured pattern. For a similar reason, single-view photometric stereo~\cite{ikehata2014photometric} results in a considerably deviated depth map. As a comparison, our approach produces a 3D point cloud that captures the major geometric features of the original object. 

\subsection{Evaluations}
\label{sec:eval}
We set the final feature length to 4, after balancing the reconstruction time and quality. For each branch of the network, we find that using a feature length that is roughly equal to twice its measurement number strikes a good balance between the feature distinctiveness and the computation time. We report $L_{\operatorname{main}}$ of networks / branches with different parameters in Tab.~\ref{table:loss}. Overall the loss decreases with the increase of the number of measurements (i.e., bandwidth). This shows that our framework can exploit the available input bandwidth to improve the feature quality. In addition, the effectiveness of adding the intensity-insensitive branch can be observed, when comparing the second column with the last, both of which have the same bandwidth at each row.
\begin{table}[htb]
  \begin{center}
   \caption{The loss $L_{\operatorname{main}}$ of networks / branches with different parameters. Each corresponding bandwidth is listed between parentheses.}
   \label{table:loss}
   \setlength{\tabcolsep}{1.1mm}{
      \begin{tabular}{c|c|c|c}
         \hline         
         \multicolumn{2}{c|}{Inten.-sensitive Branch} & \makecell[c]{Inten.-insensitive \\ Branch}  &  \makecell[c]{Our \\ Network} \\
         \hline
         56.43 (3) & 53.98 (6) & 55.32 (3) & 11.33 (6) \\
         56.85 (3) & 53.31 (8) & 53.72 (5) & 10.33 (8) \\
         53.89 (5) & 53.17 (10) & 53.72 (5) & 9.45 (10)  \\
         \hline
      \end{tabular}
   }
   \end{center}
\end{table}

\begin{figure}
   \begin{minipage}{\linewidth}
      \centering
      \includegraphics[width=0.32\linewidth]{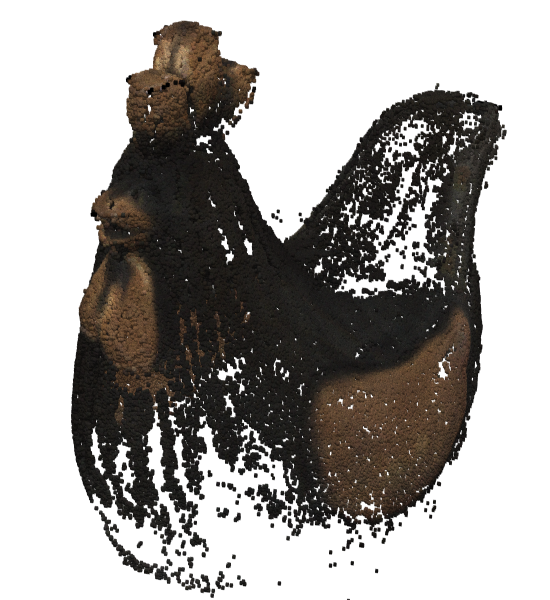}
      \includegraphics[width=0.32\linewidth]{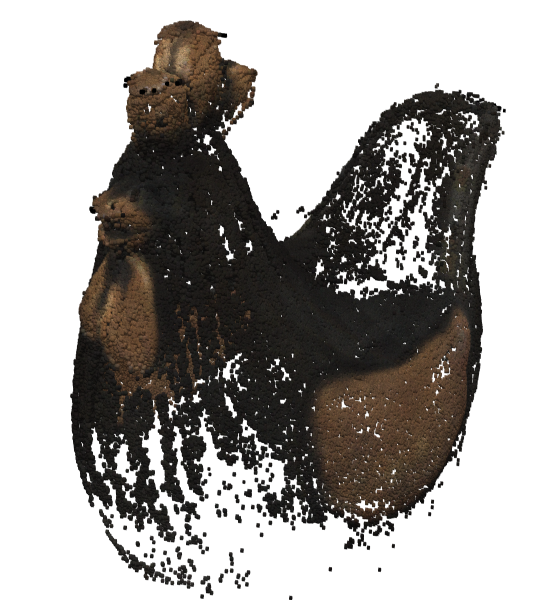}
      \includegraphics[width=0.32\linewidth]{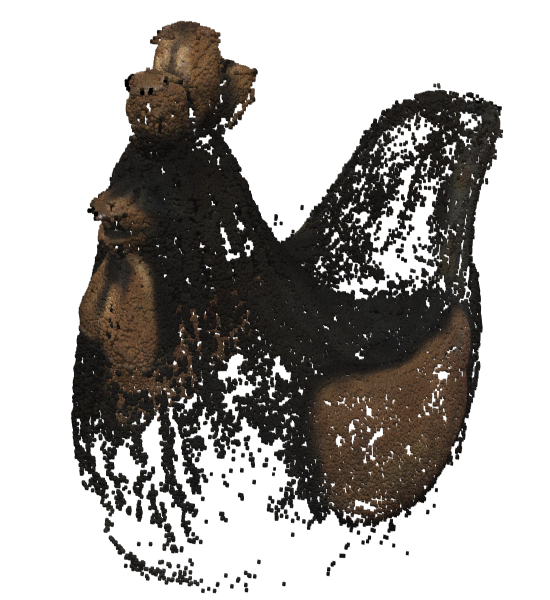}
      
         \begin{minipage}{0.32\linewidth}
            \centering
            {\small 65.1/79.8}
         \end{minipage}		
         \begin{minipage}{0.32\linewidth}
            \centering
            {\small 64.9/80.2}
         \end{minipage}		
         \begin{minipage}{0.32\linewidth}
            \centering
            {\small 64.8/77.4}
         \end{minipage}		
      
   \end{minipage}
   \caption{Impact of noise over geometric reconstruction. From the left to right, we multiply captured measurements with a a unit mean and a standard deviation of $0\%$~/~$1\%$~/~$5\%$ Gaussian noise and show the corresponding reconstructions. Quantitative errors in accuracy / completeness are reported at the bottom.}
   \label{fig:robustness}
\end{figure}

Finally, we evaluate the robustness of our features with respect to measurement noise in~\figref{fig:robustness}. The proposed framework produces reasonable results, even when a noise with 5 times the magnitude of its training counterpart is added, as shown in the figure.

\section{Conclusion}
We propose a novel framework to automatically and jointly learn to efficiently measure the physical photometric information and transform into distinctive and invariant features, which can be plugged into existing multi-view stereo pipeline for 3D reconstruction. High-quality geometric results are demonstrated over a number of daily objects with complex apperance, outperforming state-of-the-art techniques. We hope that our work will inspire interesting future work that combines multi-view and photometric cues for geometric reconstruction in a unified framework.

The proposed work is subject to a number of limitations. It does not model the appearance beyond reflectance (e.g, subsurface scattering), or explicitly handle global illumination effects like self-shadows or inter-reflections. Moreover, the current framework requires calibrated, controlled illumination. It would be promising research directions to address these limitations, to extend the idea to more general settings. We are also interested in taking  other physical dimensions (e.g., polarization / hyperspectral imaging) into considerations for the extra discriminative power.

\section*{Acknowledgement}
This work is partially supported by NSF China (61772457 \& 62022072).

{\small
\bibliographystyle{ieee_fullname}
\bibliography{learning_efficient_ps_transform_for_mvs}

\begin{thebibliography}{10}\itemsep=-1pt

\bibitem{alldrin2008photometric}
Neil Alldrin, Todd Zickler, and David Kriegman.
\newblock Photometric stereo with non-parametric and spatially-varying
  reflectance.
\newblock In {\em CVPR}, pages 1--8, 2008.

\bibitem{alldrin2007resolving}
Neil~G Alldrin, Satya~P Mallick, and David~J Kriegman.
\newblock Resolving the generalized bas-relief ambiguity by entropy
  minimization.
\newblock In {\em CVPR}, pages 1--7, 2007.

\bibitem{basri2007photometric}
Ronen Basri, David Jacobs, and Ira Kemelmacher.
\newblock Photometric stereo with general, unknown lighting.
\newblock {\em IJCV}, 72(3):239--257, 2007.

\bibitem{brown2010discriminative}
Matthew Brown, Gang Hua, and Simon Winder.
\newblock Discriminative learning of local image descriptors.
\newblock {\em TPAMI}, 33(1):43--57, 2010.

\bibitem{chen2018ps}
Guanying Chen, Kai Han, and Kwan-Yee~K Wong.
\newblock Ps-fcn: A flexible learning framework for photometric stereo.
\newblock In {\em Proceedings of the European conference on computer vision
  (ECCV)}, pages 3--18, 2018.

\bibitem{cui2017polarimetric}
Zhaopeng Cui, Jinwei Gu, Boxin Shi, Ping Tan, and Jan Kautz.
\newblock Polarimetric multi-view stereo.
\newblock In {\em CVPR}, pages 1558--1567, 2017.

\bibitem{1467495}
M. {Fiala}.
\newblock Artag, a fiducial marker system using digital techniques.
\newblock In {\em CVPR}, pages 590--596 vol. 2, 2005.

\bibitem{furukawa2015multi}
Yasutaka Furukawa and Carlos Hern{\'a}ndez.
\newblock Multi-view stereo: A tutorial.
\newblock {\em Foundations and Trends in Computer Graphics and Vision},
  9(1-2):1--148, 2015.

\bibitem{galliani2015massively}
Silvano Galliani, Katrin Lasinger, and Konrad Schindler.
\newblock Massively parallel multiview stereopsis by surface normal diffusion.
\newblock In {\em ICCV}, pages 873--881, 2015.

\bibitem{goldman2009shape}
Dan~B Goldman, Brian Curless, Aaron Hertzmann, and Steven~M Seitz.
\newblock Shape and spatially-varying brdfs from photometric stereo.
\newblock {\em TPAMI}, 32(6):1060--1071, 2009.

\bibitem{guo2020cascade}
Xiaodong Gu, Zhiwen Fan, Siyu Zhu, Zuozhuo Dai, Feitong Tan, and Ping Tan.
\newblock Cascade cost volume for high-resolution multi-view stereo and stereo
  matching.
\newblock In {\em Proceedings of the IEEE/CVF Conference on Computer Vision and
  Pattern Recognition}, pages 2495--2504, 2020.

\bibitem{hernandez2008multiview}
Carlos Hernandez, George Vogiatzis, and Roberto Cipolla.
\newblock Multiview photometric stereo.
\newblock {\em TPAMI}, 30(3):548--554, 2008.

\bibitem{ikehata2014photometric}
Satoshi Ikehata and Kiyoharu Aizawa.
\newblock Photometric stereo using constrained bivariate regression for general
  isotropic surfaces.
\newblock In {\em CVPR}, pages 2179--2186, 2014.

\bibitem{Kang:2018:ERC:3197517.3201279}
Kaizhang Kang, Zimin Chen, Jiaping Wang, Kun Zhou, and Hongzhi Wu.
\newblock Efficient reflectance capture using an autoencoder.
\newblock {\em ACM Trans. Graph.}, 37(4):127:1--127:10, July 2018.

\bibitem{Kang:2019:JOINT}
Kaizhang Kang, Cihui Xie, Chengan He, Mingqi Yi, Minyi Gu, Zimin Chen, Kun
  Zhou, and Hongzhi Wu.
\newblock Learning efficient illumination multiplexing for joint capture of
  reflectance and shape.
\newblock {\em ACM Trans. Graph.}, 38(6):165:1--165:12, Nov. 2019.

\bibitem{levoy2000digital}
Marc Levoy, Kari Pulli, Brian Curless, Szymon Rusinkiewicz, David Koller, Lucas
  Pereira, Matt Ginzton, Sean Anderson, James Davis, Jeremy Ginsberg, et~al.
\newblock The digital michelangelo project: 3d scanning of large statues.
\newblock In {\em Proc. SIGGRAPH}, pages 131--144, 2000.

\bibitem{li2020multi}
Min Li, Zhenglong Zhou, Zhe Wu, Boxin Shi, Changyu Diao, and Ping Tan.
\newblock Multi-view photometric stereo: a robust solution and benchmark
  dataset for spatially varying isotropic materials.
\newblock {\em IEEE Transactions on Image Processing}, 29:4159--4173, 2020.

\bibitem{logothetis2019differential}
Fotios Logothetis, Roberto Mecca, and Roberto Cipolla.
\newblock A differential volumetric approach to multi-view photometric stereo.
\newblock In {\em Proceedings of the IEEE/CVF International Conference on
  Computer Vision}, pages 1052--1061, 2019.

\bibitem{lu2013uncalibrated}
Feng Lu, Yasuyuki Matsushita, Imari Sato, Takahiro Okabe, and Yoichi Sato.
\newblock Uncalibrated photometric stereo for unknown isotropic reflectances.
\newblock In {\em CVPR}, pages 1490--1497, 2013.

\bibitem{nehab2005efficiently}
Diego Nehab, Szymon Rusinkiewicz, James Davis, and Ravi Ramamoorthi.
\newblock Efficiently combining positions and normals for precise 3d geometry.
\newblock {\em ACM transactions on graphics (TOG)}, 24(3):536--543, 2005.

\bibitem{NEURIPS2019_9015}
Adam Paszke, Sam Gross, Francisco Massa, Adam Lerer, James Bradbury, Gregory
  Chanan, Trevor Killeen, Zeming Lin, Natalia Gimelshein, Luca Antiga, Alban
  Desmaison, Andreas Kopf, Edward Yang, Zachary DeVito, Martin Raison, Alykhan
  Tejani, Sasank Chilamkurthy, Benoit Steiner, Lu Fang, Junjie Bai, and Soumith
  Chintala.
\newblock Pytorch: An imperative style, high-performance deep learning library.
\newblock In H. Wallach, H. Larochelle, A. Beygelzimer, F. d\textquotesingle
  Alch\'{e}-Buc, E. Fox, and R. Garnett, editors, {\em NIPS}, pages 8024--8035.
  Curran Associates, Inc., 2019.

\bibitem{pharr2016physically}
Matt Pharr, Wenzel Jakob, and Greg Humphreys.
\newblock {\em Physically based rendering: From theory to implementation}.
\newblock Morgan Kaufmann, 2016.

\bibitem{salvi2004pattern}
Joaquim Salvi, Jordi Pages, and Joan Batlle.
\newblock Pattern codification strategies in structured light systems.
\newblock {\em Pattern recognition}, 37(4):827--849, 2004.

\bibitem{schoenberger2016sfm}
Johannes~Lutz Sch\"{o}nberger and Jan-Michael Frahm.
\newblock Structure-from-motion revisited.
\newblock In {\em CVPR}, 2016.

\bibitem{schoenberger2016mvs}
Johannes~Lutz Sch\"{o}nberger, Enliang Zheng, Marc Pollefeys, and Jan-Michael
  Frahm.
\newblock Pixelwise view selection for unstructured multi-view stereo.
\newblock In {\em ECCV}, 2016.

\bibitem{schroff2015facenet}
Florian Schroff, Dmitry Kalenichenko, and James Philbin.
\newblock Facenet: A unified embedding for face recognition and clustering.
\newblock In {\em CVPR}, pages 815--823, 2015.

\bibitem{shi2012elevation}
Boxin Shi, Ping Tan, Yasuyuki Matsushita, and Katsushi Ikeuchi.
\newblock Elevation angle from reflectance monotonicity: Photometric stereo for
  general isotropic reflectances.
\newblock In {\em ECCV}, pages 455--468. Springer, 2012.

\bibitem{shi2016benchmark}
Boxin Shi, Zhe Wu, Zhipeng Mo, Dinglong Duan, Sai-Kit Yeung, and Ping Tan.
\newblock A benchmark dataset and evaluation for non-lambertian and
  uncalibrated photometric stereo.
\newblock In {\em CVPR}, pages 3707--3716, 2016.

\bibitem{EinScanP93:online}
Shining3D.
\newblock {EinScan Pro 2X Plus} handheld industrial scanner.
\newblock \url{https://www.einscan.com/handheld-3d-scanner/2x-plus/}.

\bibitem{simonyan2014learning}
Karen Simonyan, Andrea Vedaldi, and Andrew Zisserman.
\newblock Learning local feature descriptors using convex optimisation.
\newblock {\em TPAMI}, 36(8):1573--1585, 2014.

\bibitem{tian2017l2}
Yurun Tian, Bin Fan, and Fuchao Wu.
\newblock L2-net: Deep learning of discriminative patch descriptor in euclidean
  space.
\newblock In {\em CVPR}, pages 661--669, 2017.

\bibitem{vlasic2009dynamic}
Daniel Vlasic, Pieter Peers, Ilya Baran, Paul Debevec, Jovan Popovi{\'c},
  Szymon Rusinkiewicz, and Wojciech Matusik.
\newblock Dynamic shape capture using multi-view photometric stereo.
\newblock In {\em ACM SIGGRAPH Asia 2009 Papers}, pages 1--11. 2009.

\bibitem{10.5555/2383847.2383874}
Bruce Walter, Stephen~R. Marschner, Hongsong Li, and Kenneth~E. Torrance.
\newblock Microfacet models for refraction through rough surfaces.
\newblock In {\em Proceedings of the 18th Eurographics Conference on Rendering
  Techniques}, EGSR'07, page 195–206, Goslar, DEU, 2007. Eurographics
  Association.

\bibitem{woodham1980photometric}
Robert~J Woodham.
\newblock Photometric method for determining surface orientation from multiple
  images.
\newblock {\em Optical engineering}, 19(1):191139, 1980.

\bibitem{wu2017sampling}
Chao-Yuan Wu, R Manmatha, Alexander~J Smola, and Philipp Krahenbuhl.
\newblock Sampling matters in deep embedding learning.
\newblock In {\em ICCV}, pages 2840--2848, 2017.

\bibitem{wu2018specular}
Shihao Wu, Hui Huang, Tiziano Portenier, Matan Sela, Daniel Cohen-Or, Ron
  Kimmel, and Matthias Zwicker.
\newblock Specular-to-diffuse translation for multi-view reconstruction.
\newblock In {\em ECCV}, pages 183--200, 2018.

\bibitem{zagoruyko2015learning}
Sergey Zagoruyko and Nikos Komodakis.
\newblock Learning to compare image patches via convolutional neural networks.
\newblock In {\em CVPR}, pages 4353--4361, 2015.

\bibitem{zbontar2015computing}
Jure Zbontar and Yann LeCun.
\newblock Computing the stereo matching cost with a convolutional neural
  network.
\newblock In {\em CVPR}, pages 1592--1599, 2015.

\bibitem{zhou2013multi}
Zhenglong Zhou, Zhe Wu, and Ping Tan.
\newblock Multi-view photometric stereo with spatially varying isotropic
  materials.
\newblock In {\em CVPR}, pages 1482--1489, 2013.

\end{thebibliography}
}

\end{document}